%% file: draft.tex
\documentclass[preprint,3p,times,twocolumn]{elsarticle}

\journal{}

\usepackage[bookmarks,pdfborder={0 0 0},colorlinks=false,bookmarksopen=false,pdfstartview=Fit,plainpages=false,pdfpagelabels]{hyperref}
\usepackage{graphicx}
\usepackage{subfigure}
\usepackage{amssymb}
\usepackage{amsmath}
\usepackage{color}
\usepackage[svgnames,table]{xcolor}
\usepackage{comment}
\usepackage{mathrsfs}
\usepackage[ruled,lined,linesnumbered,commentsnumbered]{algorithm2e}
\usepackage{makecell}
\usepackage{booktabs}
\usepackage{multirow}
\usepackage{rotating}
\usepackage{titlesec}
\usepackage[table]{xcolor} %
\usepackage{wrapfig}
\titleformat{\subsection}{\bfseries\itshape}{\thesubsection}{1em}{} %
\titleformat{\subsubsection}{\bfseries\itshape}{\thesubsubsection}{1em}{} %

\makeatletter

\renewcommand\paragraph{\@startsection{paragraph}{4}{\z@}{3.25ex \@plus1ex \@minus.2ex}{-1em}{\normalsize\bfseries\itshape}} %

\DeclareRobustCommand\onedot{\futurelet\@let@token\@onedot}
\def\@onedot{\ifx\@let@token.\else.\null\fi\xspace}

\def\eg{\emph{e.g}\onedot} 
\def\ie{\emph{i.e}\onedot} 
 
\def\etc{\emph{etc}\onedot} \def\vs{\emph{vs}\onedot}
 
\def\etal{\emph{et al}\onedot}

\renewcommand{\arraystretch}{1.15} %

\begin{document}

\begin{frontmatter}

\title{Visual Question Answering: A Survey of Methods and Datasets}

\author{Qi Wu}
\author{Damien Teney}
\author{Peng Wang}
\author{Chunhua Shen\corref{cor1}}
\author{Anthony Dick}
\author{Anton van den Hengel}
\cortext[cor1]{Corresponding author}
\address{e-mail: firstname.lastname@adelaide.edu.au \\  School of Computer Science, The University of
Adelaide, SA 5005, Australia}

\begin{abstract}
Visual Question Answering (VQA) is a challenging task that has received increasing attention from both the computer vision and the natural language processing communities. Given an image and a question in natural language, it requires reasoning over visual elements of the image and general knowledge to infer the correct answer. In the first part of this survey, we examine the state of the art by comparing modern approaches to the problem. We classify methods by their mechanism to connect the visual and textual modalities. In particular, we examine the common approach of combining convolutional and recurrent neural networks to map images and questions to a common feature space. We also discuss memory-augmented and modular architectures that interface with structured knowledge bases. In the second part of this survey, we review the datasets available for training and evaluating VQA systems. The various datatsets contain questions at different levels of complexity, which require different capabilities and types of reasoning. We examine in depth the question/answer pairs from the Visual Genome project, and evaluate the relevance of the structured annotations of images with scene graphs for VQA. Finally, we discuss promising future directions for the field, in particular the connection to structured knowledge bases and the use of natural language processing models.
\end{abstract}


\end{frontmatter}

\tableofcontents
\clearpage

\input{introduction.tex}

\input{models.tex}
\input{models_compositional.tex}

\input{models_kb.tex}
\input{dataset.tex}
\input{dataset_real.tex}
\input{dataset_abstract.tex}

\input{dataset_kb.tex}
\input{dataset_others.tex}

\input{SGKB.tex}

\input{future.tex}

{\section*{Acknowledgements}
This research was in part supported by the Data to Decisions Cooperative Research Centre funded by the Australian Government.

\onecolumn

{
\bibliographystyle{ieee}
\bibliography{CSRef_2}
}

\end{document}

%% file: introduction.tex
\section{Introduction}

Visual question answering is a task that was proposed to connect computer vision and natural language processing (NLP), to stimulate research, and push the boundaries of both fields. On the one hand, computer vision studies methods for acquiring, processing, and understanding images. In short, its aim is to teach machines \textit{how to see}. On the the other hand, NLP is the field concerned with enabling interactions between computers and humans in natural language, \ie teaching machines \textit{how to read}, among other tasks. Both computer vision and NLP belong to the domain of artificial intelligence and they share similar methods rooted in machine learning. However, they have historically developed separately. Both fields have seen significant advances towards their respective goals in the past few decades, and the combined explosive growth of visual and textual data is pushing towards a marriage of efforts from both fields. For example, research in image captioning, \ie automatic image description \cite{donahue2014long,karpathy2014deep,mao2014deep,vinyals2014show,yao2015describing,wu2015image} has produced powerful methods for jointly learning from image and text inputs to form higher-level representations. A successful approach is to combine convolutional neural networks (CNNs), trained on object recognition, with word embeddings, trained on large text corpora.

In the most common form of Visual Question Answering (VQA), the computer is presented with an image and a textual question about this image (see examples in Figures~\ref{fig:examples1}--\ref{fig:examples3}). It must then determine the correct answer, typically a few words or a short phrase. Variants include binary (yes/no) \cite{antol2015vqa,zhang2015yin} and multiple-choice settings \cite{antol2015vqa,zhu2015visual7w}, in which candidate answers are proposed. A closely related task is to ``fill in the blank'' \cite{yu2015visual}, where an affirmation describing the image must be completed with one or several missing words. These affirmations essentially amount to questions phrased in declarative form. A major distinction between VQA and other tasks in computer vision is that the question to be answered is not determined until run time. In traditional problems such as segmentation or object detection, the single question to be answered by an algorithm is predetermined and only the input image changes. In VQA, in contrast, the form that the question will take is unknown, as is the set of operations required to answer it. In this sense, it more closely reflects the challenge of general image understanding. VQA is related to the task of textual question answering, in which the answer is to be found in a specific textual narrative (\ie reading comprehension) or in large knowledge bases (\ie information retrieval). Textual QA has been studied for a long time in the NLP community, and VQA is its extension to additional visual supporting information. The added challenge is significant, as images are much higher dimensional, and typically more noisy than pure text. Moreover, images lack the structure and grammatical rules of language, and there is no direct equivalent to the NLP tools such as syntactic parsers and regular expression matching. Finally, images capture more of the richness of the real world, whereas natural language already represents a higher level of abstraction. For example, compare the phrase `\texttt{a red hat}' with the multitude of its representations that one can picture, and in which many styles could not be described in a short sentence.

Visual question answering is a significantly more complex problem than image captioning, as it frequently requires information not present in the image. The type of this extra required information may range from common sense to encyclopedic knowledge about a specific element from the image. In this respect, VQA constitutes a truly AI-complete task \cite{antol2015vqa}, as it requires multimodal knowledge beyond a single sub-domain. This comforts the increased interest in VQA, as it provides a proxy to evaluate our progress towards AI systems capable of advanced reasoning combined with deep language and image understanding. Note that image understanding could in principle be evaluated equally well through image captioning. Practically however, VQA has the advantage of an easier evaluation metric. Answers typically contain only a few words. The long ground truth image captions are more difficult to compare with predicted ones. Although advanced evaluation metrics have been studied, this is still an open research problem \cite{li2011composing,hodosh2013framing,vedantam2014cider}.

One of the first integrations of vision and language is the ``SHRDLU'' from system from 1972 \cite{winograd1972understanding} which allowed users to use language to instruct a computer to move various objects around in a ``blocks world''. More recent attempts at creating conversational robotic agents~\cite{kollar2013toward,cantrell2010robust,matuszek2012joint,roy2003conversational} are also grounded in the visual world. However, these works were often limited to specific domains and/or on restricted language forms. In comparison, VQA specifically addresses free-form open-ended questions. The increasing interest in VQA is driven by the existence of mature techniques in both computer vision and NLP and the availability of relevant large-scale datasets. Therefore, a large body of literature on VQA has appeared over the last few years. The aim of this survey is to give a comprehensive overview of the field, covering models, datasets, and to suggest promising future directions. To the best of our knowledge, this article is the first survey in the field of VQA.

In the first part of this survey (Section~\ref{sec:models}), we present a comprehensive review of VQA methods through four categories based on the nature of their main contribution. Incremental contributions means that most methods belong to multiple of these categories (see Table~\ref{tab:methods}). First, the \textit{joint embedding approaches} (Section~\ref{subsec:Joint_e_app}) are motivated by the advances of deep neural networks in both computer vision and NLP. They use convolutional and recurrent neural networks (CNNs and RNNs) to learn embeddings of images and sentences in a common feature space. This allows one to subsequently feed them together to a classifier that predicts an answer~\cite{gao2015you,malinowski2015ask,ma2015learning}. Second, \textit{attention mechanisms} (Section~\ref{subsec:Attention_m}) improve on the above method by focusing on specific parts of the input (image and/or question). Attention in VQA \cite{zhu2015visual7w,xu2015ask,Chen2015ABC,Jiang2015compositional,andreas2015deep,yang2015stacked} was inspired by the success of similar techniques in the context of image captioning \cite{xu2015show}. The main idea is to replace holistic (image-wide) features with spatial feature maps, and to allow interactions between the question and specific regions of these maps. Third, \textit{compositional models} (Section~\ref{subsec:Compositional_m}) allow to tailor the performed computations to each problem instance. For example, Andreas \etal \cite{andreas2015deep} use a parser to decompose a given question, then build a neural network out of modules whose composition reflect the structure of the question. Fourth, \textit{knowledge base-enhanced approaches} (Section~\ref{subsec:KB_app}) address the use of external data by querying structured knowledge bases. This allows retrieving information that is not present in the common visual datasets such as ImageNet \cite{deng2009imagenet} or COCO \cite{lin2014microsoft}, which are only labeled with classes, bounding boxes, and/or captions. Information available from knowledge bases ranges from common sense to encyclopedic level, and can be accessed with no need for being available at training time \cite{wu2015ask,wang2015explicit}.

In the second part of this survey (Section~\ref{sec:dataset}), we examine datasets available for training and evaluating VQA systems. These datasets vary widely along three dimensions: (i)~their size, \ie the number of images, questions, and different concepts represented. (ii)~the amount of required reasoning, \eg whether the detection of a single object is sufficient or whether inference is required over multiple facts or concepts, and (iii)~how much information beyond that present in the actual images is necessary, be it common sense or subject-specific information. Our review points out that existing datasets lean towards visual-level questions, and require little external knowledge, with few exceptions \cite{wang2015explicit,wang2016fvqa}. These characteristics reflect the struggle with simple visual questions still faced by the current state of the art, but these characteristics must not be forgotten when VQA is presented as an AI-complete evaluation proxy. We conclude that more varied and sophisticated datasets will eventually be required.

Another significant contribution of this survey is an in-depth analysis of the question/answer pairs provided in the Visual Genome dataset (Section~\ref{sec:SGKB}). They constitute the largest VQA dataset available at the time of this writing, and, importantly, it includes rich structured images annotations in the form of scene graphs \cite{krishnavisualgenome}. We evaluate the relevance of these annotations for VQA, by comparing the occurrence of concepts involved in the provided questions, answers, and image annotations. We find out that only about 40\% of the answers directly match elements in the scene graphs. We further show that this matching rate can be significantly increased by relating scene graphs to external knowledge bases. We conclude this paper in Section~\ref{sec:FD} by discussing the potential of better connection to such knowledge bases, together with better use of existing work from the field of NLP.

%% file: models.tex
\section{Methods for VQA} %
\label{sec:models}

One of the first attempts at ``open-world'' visual question answering was proposed by Malinowski \etal \cite{malinowski2014multi}. They described a method combining semantic text parsing with image segmentation in a Bayesian formulation that samples from nearest neighbors in the training set. The method requires human-defined predicates, which are inevitably dataset-specific and difficult to scale. It is also very dependent on the accuracy of the image segmentation algorithm and of the estimated image depth information. Another early attempt at VQA by Tu \etal \cite{tu2014joint} was based on a joint parse graph from text and videos. In \cite{geman2015visual}, Geman \etal proposed an automatic ``query generator'' that is trained on annotated images and then produces a sequence of binary questions from any given test image. A common characteristic of these early approaches is to restrict questions to predefined forms. The remainder of this article focuses on modern approaches aimed at answering free-form open-ended questions. We will present methods through four categories: joint embedding approaches, attention mechanisms, compositional models, and knowledge base-enhanced approaches. As summarized in Table \ref{tab:methods}, most methods combine multiple strategies and thus belong to several categories.

\rowcolors{2}{gray!25}{} %

\begin{table*}[t]
\centering
\label{tab:methods}
\resizebox{\linewidth}{!}{
\begin{tabular}{lcccc cc}
\Xhline{2\arrayrulewidth}
~       				& Joint & Attention & Compositional & Knowledge & Answer & Image\\
Method       				& embedding & mechanism & model & base & class. / gen. & features\\
\Xhline{2\arrayrulewidth}
Neural-Image-QA \cite{malinowski2015ask}	& \checkmark      &                     &               && generation &GoogLeNet \cite{szegedy2014googlenet}                \\ %
VIS+LSTM \cite{ren2015image}       		 & \checkmark      &                     &               && classification &VGG-Net \cite{simonyan2014vgg}               \\ %
Multimodal QA \cite{gao2015you} 		& \checkmark      &                     &               && generation &GoogLeNet \cite{szegedy2014googlenet}                \\ %
DPPnet \cite{Noh2015image}  			& \checkmark      &                     &               && classification &VGG-Net \cite{simonyan2014vgg}               \\ %
MCB \cite{fukui2016multimodal}			& \checkmark      &                     &               && classification &ResNet \cite{he2015resnet}                \\ %
MCB-Att \cite{fukui2016multimodal}		& \checkmark      &   \checkmark                  &               && classification &ResNet \cite{he2015resnet}                \\ %
MRN \cite{kim2016multimodal}			& \checkmark      &   \checkmark                  &               && classification &ResNet \cite{he2015resnet}                \\ %
Multimodal-CNN \cite{ma2015learning}			& \checkmark      &                     &               && classification &VGG-Net \cite{simonyan2014vgg}               \\ %
iBOWING \cite{zhou2015simple}		& \checkmark      &                     &               && classification &GoogLeNet \cite{szegedy2014googlenet}                \\ %
VQA team \cite{antol2015vqa}		& \checkmark      &                     &               && classification &VGG-Net \cite{simonyan2014vgg}               \\ %
Bayesian \cite{kafle2016answer}    & \checkmark      &                     &               && classification &ResNet \cite{he2015resnet}                \\ %
DualNet \cite{saito2016dualnet}    & \checkmark      &                     &               && classification &VGG-Net \cite{simonyan2014vgg} \& ResNet \cite{he2015resnet}                \\ %
MLP-AQI \cite{jabri2016revisiting}    & \checkmark      &                     &               && classification &ResNet \cite{he2015resnet}                \\ %
LSTM-Att \cite{zhu2015visual7w}		& \checkmark      &    \checkmark                 &               && classification &VGG-Net \cite{simonyan2014vgg}               \\ %
Com-Mem \cite{Jiang2015compositional}		& \checkmark      &    \checkmark                 &               && generation &VGG-Net \cite{simonyan2014vgg}               \\ %
QAM \cite{Chen2015ABC}		& \checkmark      &    \checkmark                 &               && classification &VGG-Net \cite{simonyan2014vgg}               \\ %
SAN \cite{yang2015stacked}		& \checkmark      &    \checkmark                 &               && classification &GoogLeNet \cite{szegedy2014googlenet}                \\ %
SMem \cite{xu2015ask}		& \checkmark      &    \checkmark                 &               &      &classification &GoogLeNet \cite{szegedy2014googlenet}          \\ %
Region-Sel \cite{shih2015look}		& \checkmark      &    \checkmark                 &               && classification & VGG-Net \cite{simonyan2014vgg}              \\ %
FDA \cite{ilievski2016focused}		& \checkmark      &    \checkmark                 &               && classification & ResNet \cite{he2015resnet}               \\ %
HieCoAtt \cite{lu2016hierarchical}		& \checkmark      &    \checkmark                 &               && classification &ResNet \cite{he2015resnet}                \\ %
NMN \cite{andreas2015deep}	&       &    \checkmark                 &   \checkmark            && classification &VGG-Net \cite{simonyan2014vgg}               \\ %
DMN+ \cite{xiong2016}	&       &    \checkmark                 &    \checkmark           && classification &VGG-Net \cite{simonyan2014vgg}               \\ %
Joint-Loss \cite{noh2016training} &       &    \checkmark                 &    \checkmark           && classification &ResNet \cite{he2015resnet}                \\ %
Attributes-LSTM \cite{wu2015image}& \checkmark      &                    &               &     \checkmark   & generation &VGG-Net \cite{simonyan2014vgg}       \\ %
ACK \cite{wu2015ask}& \checkmark      &                    &               &     \checkmark       & generation &VGG-Net \cite{simonyan2014vgg}   \\ %
Ahab \cite{wang2015explicit}&       &                    &               &     \checkmark         & generation &VGG-Net \cite{simonyan2014vgg} \\ %
Facts-VQA \cite{wang2016fvqa}&     &                    &               &     \checkmark         & generation &VGG-Net \cite{simonyan2014vgg} \\ %
Multimodal KB \cite{zhu2015building}&     &                    &               &     \checkmark         & generation &ZeilerNet \cite{zeiler2014visualizing}  \\
\Xhline{2\arrayrulewidth}
\end{tabular}}
\caption{Overview of existing approaches to VQA, characterized by the use of a joint embedding of image and language features (Section~\ref{subsec:Joint_e_app}), the use of an attention mechanism (Section~\ref{subsec:Attention_m}), an explicitly compositional neural network architecture  (Section~\ref{subsec:Compositional_m}), and the use of information from an external structured knowledge base (Section~\ref{subsec:KB_app}). We also note whether the output answer is obtained by classification over a predefined set of common words and short phrases, or generated, typically with a recurrent neural network. The last column indicates the type of convolutional network used to obtain image feature.}
\end{table*}

\subsection{Joint embedding approaches}
\label{subsec:Joint_e_app}
\paragraph{Motivation}
The concept of jointly embedding images and text was first explored for the task of image captioning \cite{donahue2014long,karpathy2014deep,mao2014deep,vinyals2014show,yao2015describing,wu2015image}. It was motivated by the success of deep learning methods in both computer vision and NLP, which allow one to learn representations in a common feature space. In comparison to the task of image captioning, this motive is further reinforced in VQA by the need to perform further reasoning over both modalities together. A representation in a common space allows learning interactions and performing inference over the question and the image contents. Practically, image representations are obtained with convolutional neural networks (CNNs) pre-trained on object recognition. Text representations are obtained with word embeddings pre-trained on large text corpora. Word embeddings practically map words to a space in which distances reflect semantic similarities \cite{mikolov2013word2vec,pennington2014glove}. The embeddings of the individual words of a question are then typically fed to a recurrent neural network to capture syntactic patterns and handle variable-length sequences.

\begin{figure*}[t]
  \begin{center}
  \includegraphics[width=0.98\textwidth]{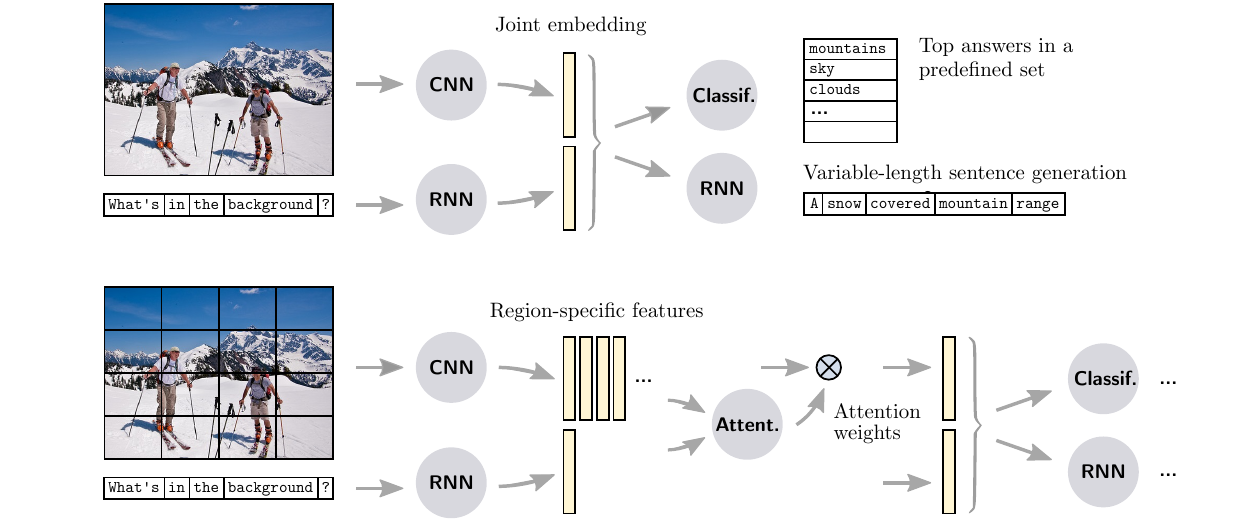}
  \end{center}
  \vspace{-5pt}
  \caption{\textit{(Top)}~A common approach to VQA is to map both the input image and question to a common embedding space (Section~\ref{subsec:Joint_e_app}). These features are produced by deep convolutional and recurrent neural networks. They are combined in an output stage, which can take the form of a classifier (\eg a multilayer perceptron) to predict short answers from predefined set or a recurrent network (\eg an LSTM) to produce variable-length phrases. \textit{(Bottom)}~Attention mechanisms build up on this basic approach with a spatial selection of image features. Attention weights are derived from both the image and the question and allow the output stage to focus on relevant parts of the image.}
  \label{fig:jointEmb}
\end{figure*}

\paragraph{Methods}
Malinowski \etal \cite{malinowski2015ask} propose an approach named ``Neural-Image-QA'' with a Recurrent Neural Network (RNN) implemented with Long Short-Term Memory cells (LSTMs) (Figure~\ref{fig:jointEmb}). The motivation behind RNNs is to handle inputs (questions) and outputs (answers) of variable size. Image features are produced by a CNN pre-trained for object recognition. Question and image features are both fed together to a first ``encoder'' LSTM. It produces a feature vector of fixed-size that is then passed to a second ``decoder'' LSTM. The decoder produces variable-length answers, one word per recurrent iteration. At each iteration, the last predicted word is fed through the recurrent loop into the LSTM until a special \texttt{<END>} symbol is predicted. Several variants of this approach were proposed. For example, the ``VIS+LSTM'' of Ren \etal \cite{ren2015image} directly feed the feature vector produced by the encoder LSTM into a classifier to produce single-word answers from a predefined vocabulary. In other words, they formulate the answering as a classification problem, whereas Malinowski \etal \cite{malinowski2015ask} was treating it as a sequence generation procedure. Ren \etal \cite{ren2015image} propose other technical improvements with the ``2-VIS+BLSTM'' model. It uses two sources of image features as input, fed to the LSTM at the start and at the end of the question sentence. It also uses LSTMs that scan questions in both forward and backward directions. Those bidirectional LSTMs better capture relations between distant words in the question.

Gao \etal~\cite{gao2015you} propose a slightly different method named ``Multimodal QA'' (mQA). It employs LSTMs to encode the question and produce the answer, with two differences from \cite{malinowski2015ask}. First, whereas \cite{malinowski2015ask} used common shared weights between the encoder and decoder LSTMs, mQA learns distinct parameters and only shares the word embedding. This is motivated by potentially different properties (\eg in terms of grammar) of questions and answers. Second, the CNN features used as image representations are not fed into the encoder prior to the question, but at every time step.

Noh \etal \cite{Noh2015image} tackle VQA by learning a CNN with a dynamic parameter layer (DPPnet) of which the weights are determined adaptively based on the question. For the adaptive parameter prediction, they employ a separate parameter prediction network, which consists of gated recurrent units (GRUs, a variant of LSTMs) taking a question as input and producing candidate weights through a fully-connected layer at its output. This arrangement was shown to significantly improve answering accuracy compared to \cite{malinowski2015ask,ren2015image}. One can note a similarity in spirit with the modular approaches of Section~\ref{subsec:Compositional_m}, in the sense that the question is used to tailor the main computations to each particular instance.

Fukui \etal \cite{fukui2016multimodal} propose a pooling method to perform the joint embedding visual and text features. They perform their ``Multimodal Compact Bilinear pooling'' (MCB) by randomly projecting the image and text features to a higher-dimensional space and then convolve both vectors with multiplications in the Fourier space for efficiency. Kim \etal~\cite{kim2016multimodal} use a multimodal residual learning framework (MRN) to learn the joint representation of images and language. Saito \etal \cite{saito2016dualnet} propose a ``DualNet'' which integrates two kinds of operations, namely element-wise summations and element-wise multiplications to embed their visual and textual features. Similarly as \cite{ren2015image, Noh2015image}, they formulate the answering as a classification problem over a predefined set of possible answers. Kafle \etal \cite{kafle2016answer} integrate an explicit prediction of the type of expected answer from the question and formulate the answering in a Bayesian framework.

Some other works do not make use of RNNs to encode questions. Ma \etal \cite{ma2015learning} use CNNs to process the questions. Features from the image CNN and the text CNN are embedded in a common space through additional layers (a ``multimodal CNN'') forming an overall homogeneous convolutional architecture. Zhou \etal \cite{zhou2015simple} and Antol \etal \cite{antol2015vqa} both use a traditional bag-of-words representation of the questions.

\paragraph{Performance and limitations}
We summarize performances of all discussed methods and datasets in Tables~\ref{tab:DAQUAR_All}--\ref{tab:VQA_result}. The ``Neural-Image-QA'', as one of the earliest introduced methods, is considered the \textit{de facto} baseline result. The ``2-VIS+BLSTM'' improves slightly on the DAQUAR dataset, mostly thanks to the bidirectional LSTM used to encode the questions. The ``mQA'' model was unfortunately not tested on publicly available datasets and is therefore not comparable. The DPPnet \cite{Noh2015image} showed significant benefit from tailoring computations adaptively to each question through the dynamic parameter layer. At the time of its publication, it outperformed other joint embeddings methods \cite{malinowski2015ask,ren2015image,antol2015vqa,zhou2015simple}. The more recent MCB pooling \cite{fukui2016multimodal} and multimodal residual learning (MRN) bring further improvements and achieve the top performances at the time of this writing.

The joint embedding approaches are straightforward in their principle and constitute the base of most current approaches to VQA. The latest improvements, exemplified by MCB and MRN, still showed potential room for improvement on both the extraction of features and their projection to the embedding space.

\subsection{Attention mechanisms}
\label{subsec:Attention_m}
\paragraph{Motivation}
A limitation of most models presented above is to use global (image-wide) features to represent the visual input. This may feed irrelevant or noisy information to the prediction stage. The aim of attention mechanisms is to address this issue by using local image features, and allowing the model to assign different importance to features from different regions. An early application of attention to visual tasks was proposed in the context of image captioning by Xu \etal \cite{xu2015show}. The attentional component of their model identifies salient regions in an image, and further processing then focuses the caption generation on those regions. This concept translates readily to the task of VQA for focusing on image regions relevant to the question. In some respect, the attention process forces an explicit additional step in the reasoning process that identifies ``where to look'' before performing further computations.

Although attention models were inspired by computational models of human vision, the apparent resemblance with biological systems can be misleading. Attention in artificial neural networks likely helps by allowing additional non-linearities and/or types of interactions (\eg multiplicative interaction through attention weights), whereas attention in biological visual systems is more likely a consequence of limited resources such as resolution, field of view, and processing capacity.

\paragraph{Methods}
Zhu \etal \cite{zhu2015visual7w} described how to add spatial attention to the standard LSTM model. The computations performed by an attention-enhanced LSTM are described as follows:
\begin{eqnarray}
  i_t &=& \sigma(W_{xi}x_t+W_{hi}h_{t-1}+W_{zi}z_t+b_i) \\
  f_t &=& \sigma(W_{xf}x_t+W_{hf}h_{t-1}+W_{zf}z_t+b_f) \\
  o_t &=& \sigma(W_{xo}x_t+W_{ho}h_{t-1}+W_{zo}z_t+b_o) \\
  g_t &=& \tanh(W_{xg}x_t+W_{hg}h_{t-1}+W_{zg}z_t+b_c) \\
  c_t &=& f_t\odot c_{t-1}+i_t\odot g_t \\
  h_t &=& o_t \odot \tanh(c_t)
\end{eqnarray}
where $\sigma$ is a sigmoid nonlinearity, and $i_t, f_t, c_t, o_t$ are the input, forget, memory, output state of the LSTM. The various $W$ and $b$ matrices are trained parameters and $\odot$ represents element-wise products with gate values. $x_t$ is the input (\eg a word of a question) and $h_t$ is the hidden state at time step $t$. The attention mechanism is introduced by the term $z_t$ , which is a weighted average of convolutional features that depends upon the previous hidden state and the convolutional features:
\begin{eqnarray}
  e_t &=& w_a^T\tanh(W_{he}h_{t-1}+W_{ce}C(I))+b_a \\
  a_t &=& {\rm softmax}(e_t)\\
  z_t &=& a_t^T C(I)
\end{eqnarray}
where $C(I)$ represents the convolutional feature map of image $I$. The \textit{attention term} $a_t$ sets the contribution of each convolutional feature at the $t$-th step. Large values in $a_t$ indicate more relevance of the corresponding region to the question. In this formulation, a standard LSTM can be considered as a special case with values in $a_t$ set uniformly, \ie each region contributing equally. A similar mechanism as above was employed by Jiang \etal \cite{Jiang2015compositional}.

Chen \etal \cite {Chen2015ABC} use a mechanism different from the above word-guided attention. They generate a ``question-guided attention map'' (QAM) by searching for visual features that correspond to the semantics of the input question in the spatial image feature map. The search is achieved by convolving the visual feature map with a configurable convolutional kernel. This kernel is generated by transforming the question embeddings from the semantic space into the visual space, which contains the visual information determined by the intent of the question. Yang \etal \cite{yang2015stacked} also employ this scheme with ``stacked attention networks'' (SAN) that infer the answer iteratively. Xu \etal \cite{xu2015ask} propose a ``multi-hop image attention scheme'' (SMem). The first hop is a word-guided attention, while a second hop is question-guided. In \cite{shih2015look}, the authors generate image regions with object proposals and then select regions relevant to the question and possible answer choices. Similarly, Ilievski \etal \cite{ilievski2016focused} employ off-the-shelf object detectors to identify regions related to the key words of the question and then fuse information from those regions with global features with an LSTM. Lu \etal \cite{lu2016hierarchical} present a ``hierarchical co-attention model'' (HieCoAtt) that jointly reasons about image and question attention. Whereas the works described above focus only on visual attention, HieCoAtt processes image and question symmetrically, in the sense that the image representation guides attention over the question and vice versa. Most recently, Fukui \etal \cite{fukui2016multimodal} combine the attention mechanism into their ``Multimodal Compact Bilinear pooling'' (MCB) already mentioned in Section~\ref{subsec:Joint_e_app}.

Andreas \etal \cite{andreas2015deep} employ attention mechanisms in a different manner. They propose a compositional model that builds a neural network from modules tailored to each question, as described in more details in Section~\ref{subsubsec:modules}. Most of these modules operate in the space of attentions, either producing an attention map from an image (\ie identifying a salient object), performing unary operations (\eg inverting the attention from an object to the context around it), or interactions between attentions (\eg a subtracting an attention map from another).

\paragraph{Performance and limitations}
The reported uses of attention mechanisms always improve over models that use global image features. For example, the authors in \cite{zhu2015visual7w} show that the attention-enhanced LSTM described above outperforms the ``VIS+LSTM''  model \cite{ren2015image} in both ``Telling'' and ``Grounding'' tasks of the `Visual7W' dataset (see Section~\ref{subsec:datasetreal}). The multiple attention layers of SAN \cite{yang2015stacked} bring further improvements over only one layer of attention \cite{Jiang2015compositional,Chen2015ABC}, especially on the VQA dataset. The HieCoAtt model \cite{lu2016hierarchical} shows benefit from the hierarchical representation of the question and also from the co-attention mechanism (question-guided visual attention and image-guided question attention).

Interestingly, attention mechanisms improve the overall accuracy on all VQA datasets, but closer inspection by question type show little or no benefit on binary (yes/no) questions. One hypothesis is that binary questions typically require longer chains of reasoning, whereas open-ended questions often require identifying and naming only one concept from the image. Therefore, improving on binary questions will likely require other innovations than visual attention. The output in end-to-end joint embedding approaches --~regardless of the use of attention~-- is produced by a simple mapping from the co-embedded visual and textual features to the answer, learned over a large number of training examples. Little insight is available as to how an output answer arises. It can be debated whether any ``reasoning'' is performed and/or encoded in the mapping. Another important issue is raised by asking whether questions can be answered from the given visual input alone. Oftentimes, they require prior knowledge ranging common sense to subject-specific and even expert-level knowledge. How such information can be provided to VQA systems and incorporated into the reasoning is still an open question (see Section~\ref{subsec:KB_app}).

%% file: models_compositional.tex
\subsection{Compositional Models}
\label{subsec:Compositional_m}

The methods discussed so far present limitations related to the monolithic nature of the CNNs and RNNs used to extract representations of images and sentences. An increasingly popular research direction in the design of artificial neural networks is to consider modular architectures. This approach involves connecting distinct modules designed for specific desired capabilities such as memory or specific types of reasoning. A potential advantage is a better use of supervision. On the one hand, it facilitates transfer learning, since a same module can be used and trained within different overall architectures and tasks (see Section \ref{subsubsec:modules}). On the other hand, it allows to use ``deep supervision'', \ie optimizing an objective that depends on the outputs of internal modules (\eg which supporting facts an attention mechanism should focus on \cite{weston2014}). Other models discussed in Section~\ref{subsec:Attention_m} (attention models) and Section~\ref{subsec:KB_app} (connections to knowledge bases) also fall in the category of modular architectures. We focus here on two specific models whose main contribution is in the modular aspect, namely the Neural Module Networks (NMN) and the Dynamic Memory Networks (DMN).

\subsubsection{Neural Module Networks}
\label{subsubsec:modules}

\begin{figure}[t]
  \begin{center}
  \includegraphics[width=0.99\columnwidth]{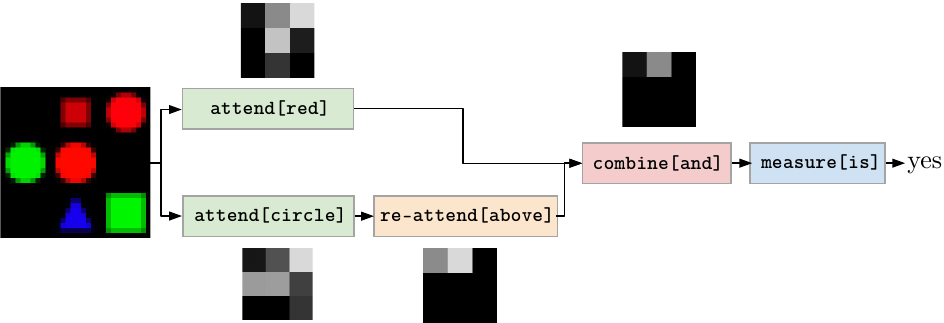}
  \end{center}
  \vspace{-5pt}
  \caption{The Neural Module Networks (NMN, Section~\ref{subsubsec:modules}) leverage the compositional structure of questions, \eg here ``Is there a red shape above a circle ?'' from the \textit{Shapes} dataset (Section~\ref{subsubsect:shapes}). The parsing of the question leads to assembling modules that operate in the space of attentions. Two \texttt{attend} modules locate red shapes and circles, \texttt{re-attend[above]} shifts the attention above the circles, \texttt{combine} computes their intersection, and \texttt{measure[is]} inspects the final attention and determines that it is non-empty (figure adapted from \cite{andreas2015deep}).}
  \label{fig:compositional}
\end{figure}

\paragraph{Motivation}
The Neural Module Networks (NMN) are introduced by Andreas \etal. in \cite{andreas2015deep} and extended in \cite{andreas2016learning}. They are specifically designed for VQA, with the intention of exploiting the compositional linguistic structure of the questions. Questions vary greatly in the level of complexity. For example, \textit{Is this a truck ?} only requires retrieving one piece of information from the image, whereas \textit{How many objects are to the left of the toaster ?} requires multiple processing steps, such as recognition and counting. NMNs reflect the complexity of a question in a network that is assembled on-the-fly for each instance of the problem. The tactic is related to approaches in textual QA \cite{liang2013dependencybased} that use semantic parsers to turn questions into logical expressions. A significant contribution of NMNs is to apply this logical reasoning over continuous visual features, instead of discrete or logical predicates.

\paragraph{Method}
The method is based on a semantic parsing of the question using a well-known tool in the NLP community. The parse tree is turned into an assembly of modules from a predefined set, which are then used together to answer the question. Crucially, all modules are independent and composable (Figure~\ref{fig:compositional}). In other words, the computation performed will be different for each problem instance, and a problem instance at test time may use a set of modules that were not seen together during training.

The inputs and outputs of the modules can be of three types: image features, attentions (regions) over images, and labels (classification decisions). A set of possible modules is predefined, each by its type of input and output, but their exact behavior will be acquired through end-to-end training on specific problem instances. The training therefore does not need additional supervision than triples of images, questions, and answers.

The parsing of the question is a crucial step, which is performed with the Standford dependency parser \cite{demarneffe2008parser} which basically identifies grammatical relations between parts of the sentence. The authors of the NMNs then use ad hoc hand-written rules to deterministically transform parse trees into structured queries, in the form of compositions of modules \cite{andreas2015deep}. In their second paper \cite{andreas2016learning}, they additionally learn a ranking function to select the best structures from candidate parses. The whole procedure still uses strong simplifying assumptions about language in the question. The visual features are provided by a fixed, pre-trained VGG CNN \cite{simonyan2014vgg}.

\paragraph{Performance and limitations}
The Neural Module Networks were evaluated on the VQA benchmark, and shows different strengths and weaknesses than competing approaches. It generally outperforms competitors on questions with a compositional structure, \eg requiring an object to be located and one of its attributes described. However, many of questions in the VQA dataset are quite simple, and require little composition or reasoning. The authors introduced a new dataset, named ``Shapes'', of synthetic images (Section \ref{subsubsect:shapes}) paired with complex questions involving spatial relations, set-theoretic reasoning, and shape and attribute recognition.

The limitations of the method are inherent to the bottleneck formed during the parsing of the question. This stage fixes the network structure and errors cannot be recovered from. Moreover, the assembly of modules uses aggressive simplification of the questions that discards some grammatical cues. As a workaround, the authors obtain the final answer by averaging their output with the one from a classical LSTM question encoder.

The potential of the NMNs is dimmed in practice by the lack of truly complex questions in the VQA benchmark. The results reported on this dataset use a restricted subset of possbible modules, presumably to avoid over-fitting. Results on the synthetic Shapes dataset show that semantic structure prediction does improve generalization in deep networks. The overall approach presents the potential of addressing the combinatorial explosion of concepts and relations that can arise in open-world VQA. Finally, note that the general formulation of NMNs can encompass other approaches, including the memory networks presented below, which may be formulated as a composition of ``attention'' and ``classifier'' modules.

\subsubsection{Dynamic Memory Networks}

\paragraph{Motivation}

The Dynamic Memory Networks (DMN) are neural networks with a particular modular architecture. They were described in \cite{kumar2015}, with a number of variants proposed concurrently \cite{weston2014,sukhbaatar2015,bordes2015,peng2015reasoning}. Most of these were applied to textual QA. We focus here on the work of Xiong \etal \cite{xiong2016}, who adapted them to VQA. DMNs fall into the broader category of memory-augmented networks, which perform read and write operations on an internal representation of the input. This mechanism, similarly to attention (see Section \ref{subsec:Attention_m}), is designed to address tasks that require complex logical reasoning by modeling interaction between multiple parts of the data over several passes.

\paragraph{Method}

The dynamic memory networks are composed of 4 main modules \cite{kumar2015} that allow independence in their particular implementation. The input module transforms the input data into a set of vectors called  ``facts''. Its implementation (described below) varies depending on the type of input data. The question module computes a vector representation of the question, using a gated recurrent unit (GRU, a variant of LSTM). The episodic memory module retrieves the facts required to answer the question. They key is to allow the episodic memory module to pass multiple times over the facts to allow transitive reasoning. It incorporates an attention mechanism that selects relevant facts and an update mechanism that generates a new memory representation from interactions between its current state and the retrieved facts. The first state is initialized with the representation from the question module. Finally, the answer module uses the final state of the memory and the question to predict the output with a multinomial classification for single words, or a GRU for datasets where a longer sentence is required.

The input module for VQA \cite{xiong2016} extracts image features with a VGG CNN \cite{simonyan2014vgg} over small image patches. These features are fed to a GRU in the manner of the words of a sentence, traversing the image in a snake-like fashion. This is an ad hoc adaptation of the original input module in \cite{kumar2015} that used a GRU to process words of sentences. The episodic memory module also includes an attention mechanism to focus on particular image regions.

Let us also mention here the work of Noh \etal \cite{noh2016training}. Their approach has similarities with memory networks in the use of an internal memory-like unit, over which multiple passes are performed. The main novelty is the use of a loss over each of these passes, instead of a single one on the final results. After training, the inference at test time is performed using only one such pass.

\paragraph{Performance and limitations}

The dynamic memory networks were evaluated on the DAQUAR and VQA benchmarks, and show competitive performance for all types of questions. Compared to Neural Module Networks (Section \ref{subsubsec:modules}), they perform similarly on yes/no questions, slightly worse on numerical questions, but markedly better on all other types of questions. The issue with counting likely arises from the limited granularity of the fixed image patches, which may cross object boundaries.

Interestingly, the paper presents competitive results on both VQA and text-based QA \cite{xiong2016} using essentially a same method, except for the input module. The text QA dataset used \cite{weston2015babi} requires inference over multiple facts, which is a positive indicator of the reasoning capabilities of this model. A potential criticism for applying a same model to text and images stems from the intrinsically different nature of sequences of words, and sequences of image patches. The temporal dimension of a textual narrative is different than relative geometrical positions, although both seem to be handled adequately by GRUs in practice.

%% file: models_kb.tex
\subsection{Models using external knowledge bases}
\label{subsec:KB_app}

\paragraph{Motivation}
The task of VQA involves understanding the contents of images, but often requires prior non-visual, information, which can range from ``common sense'' to topic-specific or even encyclopedic knowledge. For example, to answer the question ``How many mammals appear in this image ?'', one must understand the word ``mammal'' and know which animals belong to this category. This observation allows pinpointing two major weaknesses of the joint embedding approaches (Section~\ref{subsec:Joint_e_app}). First, they can only capture knowledge that is present in the training set, and it is obvious than efforts at scaling up datasets will never reach a complete coverage of the real world. Second, the neural networks trained in such approaches have a limited capacity, which is also inevitably deemed to be surpassed by the amount of information we wish to learn.

An alternative is to decouple the reasoning (\eg as a neural network) from the actual storage of data or knowledge. A substantial amount of research has been devoted to structured representations of knowledge. This led to the development of large-scale Knowledge Bases (KB) such as
DBpedia~\cite{auer2007dbpedia}, 
Freebase~\cite{bollacker2008freebase},
YAGO~\cite{hoffart2013yago2,mahdisoltani2014yago3},
OpenIE~\cite{banko2007open,etzioni2011open,fader2011identifying},
NELL~\cite{carlson2010toward},
WebChild~\cite{tandon2014acquiring,tandon2014webchild},
and ConceptNet~\cite{liu2004conceptnet}.
These databases store common sense and factual knowledge in a machine readable fashion. Each piece of knowledge, referred to as a fact, is typically represented as a triple \texttt{(arg1,rel,arg2)}, where \texttt{arg1} and \texttt{arg2} represent two concepts and \texttt{rel} represents a relationship between them. The collection of such facts forms a interlinked graph, which is often described according to a Resource Description Framework (RDF~\cite{rdf2014resource}) specification and can be accessed by query languages such as SPARQL~\cite{prud2008sparql}. Linking such knowledge bases to VQA methods allows separating the reasoning from the representation of prior knowledge in a practical and scalable manner.

\paragraph{Method}
Wang \etal~\cite{wang2015explicit} propose a VQA framework named ``Ahab'' that uses DBpedia, one of the largest structured knowledge bases. Visual concepts are first extracted from the given image with CNNs, and they are then associated with nodes from DBpedia that represent similar concepts. Whereas the joint embedding approaches (Section~\ref{subsec:Joint_e_app}) learn a mapping from images/questions to answers, the authors propose here to learn a mapping images/questions to \emph{queries} over the constructed knowledge graph. The final answer is obtained by summarizing the results of this query. The main limitation of \cite{wang2015explicit} is to handle limited types of questions. Although the questions can be provided in natual language, they are parsed using manually designed templates. An improved method named FVQA \cite{wang2016fvqa} uses an LSTM and a data-driven approach to learn the mapping of images/questions to queries. This work also uses two additional knowledge bases, ConceptNet and WebChild.

An interesting byproduct of the explicit representation of knowledge is that the above methods can indicate how they arrived to the answer by providing the chain of reasoning~\cite{wang2015explicit} or the supporting facts~\cite{wang2016fvqa} used in the inference process. This contrasts with monolithic neural networks which provide little insight into the computations performed to produce their final answer.

Wu \etal~\cite{wu2015ask} proposed a joint embedding approach that also benefits from external knowledge bases. Given an image, they first extract semantic attributes with a CNN. External knowledge related to these attributes is then retrieved from a version of DBpedia containing short descriptions, which are embedded into fixed-size vectors with Doc2Vec. The embedded vectors are fed into an LSTM model that interprets them with the question and finally generates an answer. This method still learns a mapping from questions to answers (as other joint embedding methods) and cannot provide information about the reasoning process.

\paragraph{Performance and limitations}
Both Ahab and FVQA focus specifically on visual questions requiring external knowledge. Most existing VQA datasets include a majority of questions that require little amount of prior knowledge, and performance on these datasets thus poorly reflect the particular capabilities of these methods. The authors of those two methods thus include evaluation on new small-scale datasets (see Section~\ref{subsec:datasetkb}). Ahab \cite{wang2015explicit} significantly outperforms joint embedding approaches on its KB-VQA dataset~\cite{wang2015explicit} in terms of overall accuracy ($69.6\%$ \vs $44.5\%$). In particular, Ahab becomes significantly better than joint embedding approaches on visual questions requiring a higher level of knowledge. Similarly, the FVQA approach~\cite{wang2016fvqa} also performs much better than conventional approaches~\cite{wang2016fvqa} in terms of overall top-1 accuracy ($58.19\%$ \vs $23.37\%$). An issue in the evaluation of both of these methods is the limited number of question types and the small scale of the datasets.

The approach of Wu \etal. \cite{wu2015ask} is evaluated on the Toronto COCO-QA and VQA datasets, and shows an advantage in using the external KB in terms of average accuracy.

%% file: dataset.tex
\section{Datasets and evaluation}
\label{sec:dataset}
\label{subsec:datasets}

A number of datasets have been proposed specifically for research on VQA. They contain, at the minimum, triples made of an image, a question, and its correct answer. Additional annotations are sometimes provided, such as image captions, image regions supporting the answers, or multiple-choice candidate answers. Datasets and questions within the datasets vary widely in their complexity, the amount of reasoning and of non-visual (\eg ``common sense'') information required to infer the correct answer. This section contains a comprehensive comparison of the available datasets and discusses their suitability for evaluating different aspects of VQA systems. We broadly classify dataset according to their type of images (natural, clipart, synthetic). Key characteristics are summarized in Table~\ref{tab:datasets}. See Figures~\ref{fig:examples1}--\ref{fig:examples3} for examples from various datasets.

A given dataset is typically used for both training and evaluating a VQA system. The open-ended nature of the task suggests however that other, large-scale sources of information would be beneficial and likely necessary train practical VQA systems. Some datasets specifically address this aspect through annotations of supporting facts in structured non-visual knowledge bases (Section~\ref{subsec:datasetkb}).

Other datasets non-specific to VQA are worth mentioning. They target other tasks involving vision and language, such as image captioning (\eg \cite{chen2015microsoft,hodosh2013framing,lin2014microsoft,young2014image}), generating~\cite{mao2015generation} and understanding~\cite{kazemzadeh2014referit,hu2015natural} referring expressions for retrieval of images and objects in natural language. Those datasets go beyond the scope of this article (see \cite{ferraro2015datasets} for a review), but are a potential source of additional training data for VQA since they combine images with textual annotations.

%% file: dataset_real.tex
\subsection{Datasets of natural images}
\label{subsec:datasetreal}

\begin{sidewaystable}
 \label{tab:datasets}
 \centering
 \resizebox{\linewidth}{!}{
 \rowcolors{3}{gray!25}{} %
 \begin{tabular}{lccccccccc}
  \hline
  ~			& Source of & Number of & Number of	& Num. questions /	& Num. question & Question		& Average quest.	& Average ans.	& Evaluation \\
  Dataset	& images	& images	& questions	& num. images		& categories	& collection	& length			& length		& metrics \\ \hline
  DAQUAR \cite{malinowski2014multi}  &  NYU-Depth V2  &  1,449  &  12,468  &  8.6  &  4  &  Human  &  11.5  &  1.2  &  Acc. \&  WUPS \\
  COCO-QA \cite{ren2015image}  &  COCO  &  117,684  &  117,684  &  1.0  &  4  &  Automatic  &  8.6  &  1.0  &  Acc. \&  WUPS \\
  FM-IQA \cite{gao2015you}  &  COCO  &  120,360  &  -  &  -  &  -  &  Human  &  -  &  -  &  Human \\
  VQA-real \cite{antol2015vqa}  &  COCO  &  204,721  &  614,163  &  3.0  &  20+  &  Human  &  6.2  &  1.1  &   Acc. against 10 humans \\
  Visual Genome \cite{krishnavisualgenome}  &  COCO  &  108,000  &  1,445,322  &  13.4  &  7  &  Human  &  5.7  &  1.8  &  Acc. \\
  Visual7W \cite{zhu2015visual7w}  &  COCO  &  47,300  &  327,939  &  6.9  &  7  &  Human  &  6.9  &  1.1  &  Acc. \\
  Visual Madlibs \cite{yu2015visual}  &  COCO  &  10,738  &  360,001  &  33.5  &  12  &  Human  &  6.9  &  2.0  &   Acc.\\
  VQA-abstract \cite{antol2015vqa}  &  Clipart &  50,000  &  150,000  &  3.0  &  20+  &  Human  & 6.2   &  1.1  &  Acc. \\
  VQA-balanced \cite{zhang2015yin}  &  Clipart &  15,623  &  33,379  &  2.1  &  1  &  Human  &  6.2  &  1.0  &  Acc. \\
  KB-VQA \cite{wang2015explicit}  &  COCO  &  700  &  2,402  &  3.4  &  23  &  Human  & 6.8   & 2.0   &  Human \\
  FVQA \cite{wu2016ijcv}  &  COCO \& ImageNet  &  1,906  &  4,608  &  2.5  &  12  &  Human  &  9.7  & 1.2   &  Acc. \\
  \hline
 \end{tabular}}
 \caption{Major datasets for VQA and their main characteristics. See Section~\ref{subsec:datasetreal} for discussion.}
\end{sidewaystable}

An early effort at compiling a dataset specifically for VQA was presented by Geman \etal \cite{geman2015visual}. The dataset comprises questions generated from templates from a fixed vocabulary of objects, attributes, and relationships between objects. Another early dataset was presented in \cite{tu2014joint} by Tu \etal. They study the joint parsing of videos and text to answer queries, and consider two datasets containing 15 video clips each. These two examples are restricted to limited settings and are of relatively small size. We discuss below the open-world large-scale datasets in use today.

\paragraph{DAQUAR} 
The first VQA dataset designed as benchmark is the DAQUAR, for DAtaset for QUestion Answering on Real-world images \cite{malinowski2014multi}. It was built with images from the NYU-Depth v2 dataset \cite{silberman2012indoor}, which contains 1449 RGBD images of indoor scenes, together with annotated semantic segmentations. The images of DAQUAR are split to 795 training and 654 test images. Two types of question/answer pairs are collected. First, \textit{synthetic questions/answers} are generated automatically using 8 predefined templates and the existing annotations of the NYU dataset. Second, \textit{human questions/answers} are collected from 5 annotators. They were instructed to focus on basic colors, numbers, objects (894 categories), and sets of those. Overall, 12,468 question/answer pairs were collected, of which 6,794 are to be used for training and 5,674 for testing. The large size of DAQUAR was key to enable the development and training of the early methods for VQA with deep neural networks \cite{malinowski2015ask,ren2015image,ma2015learning}. The main disadvantage of DAQUAR is the restriction of answers to a predefined set of 16 colors and 894 object categories. The dataset also presents strong biases showing that humans tend to focus on a few prominent objects, such as tables and chairs \cite{malinowski2014multi}.

\paragraph{COCO-QA} 
The COCO-QA dataset \cite{ren2015image} represents a substantial effort to increase the scale of training data for VQA. This dataset uses images from the Microsoft Common Objects in Context data (COCO) dataset \cite{lin2014microsoft}. COCO-QA includes 123,287 images (72,783 for training and 38,948 for testing) and each image has one question/answer pair. They were automatically generated by turning the image descriptions part of the original COCO dataset into question/answer form. The questions are categorized into four types based on the type of expected answer: object, number, color, and location. A side-effect of the automatic conversion of captions is a high repetition rate of the questions. Among the 38,948 questions of the test set, 9,072 (23.29\%) of them also appear as training questions.

\paragraph{FM-IQA} 
The FM-IQA (Freestyle Multilingual Image Question Answering) dataset \cite{gao2015you} uses 123,287 images, also sourced from the COCO dataset. The difference with COCO-QA is that the questions/answers are provided here by humans through the Amazon Mechanical Turk crowd-sourcing platform. The annotators were free to give any type of questions, as long as they relate to the contents of each given image. This lead to a much greater diversity of questions than in previously-available datasets. Answering the questions typically requires both understanding the visual contents of the image and incorporating prior ``common sense'' information. The dataset contains 120,360 images and 250,560 question/answer pairs, which were originally provided in Chinese, then converted into English by human translators.

\paragraph{VQA-real}
One of the most widely used dataset comes from the VQA team at Virginia Tech, commonly referred to simply as VQA \cite{antol2015vqa}. It comprises two parts, one using natural images named VQA-real, and a second one with cartoon images named VQA-abstract, which we will discuss in Section~\ref{subsec:datasetclip}. VQA-real comprises 123,287 training and 81,434 test images, respectively, sourced from COCO \cite{lin2014microsoft}. Human annotators were encouraged to provide interesting and diverse questions. In contrast to the datasets mentioned above, binary (\ie yes/no) questions were also allowed. The dataset also allows evaluation in a multiple-choice setting, by providing 17 additional (incorrect) candidate answers for each question. Overall, it contains 614,163 questions, each having 10 answers from 10 different annotators. The authors performed a very detailed analysis of the dataset \cite{antol2015vqa} in terms of questions types, question/answer lengths, \etc. They also conducted a study to investigate whether questions required prior non-visual knowledge, judged by polling humans. A majority of subjects (at least 6 out of 10) estimated that common sense was required for 18\% of the questions. Only 5.5\% of the questions were estimated to require adult-level knowledge. These modest figures show that little more than purely visual information is required to answer most questions.

\paragraph{Visual Genome and Visual7W} 
The Visual Genome QA dataset \cite{krishnavisualgenome} is, at the time of this writing, the largest available dataset for VQA, with 1.7 million question/answer pairs. It is built with images from the Visual Genome project \cite{krishnavisualgenome}, which includes unique structured annotations of scene contents in the form of scene graphs. These scene graphs describe the visual elements of the scenes, their attributes, and relationships between them. Human subjects provided questions that must start with one of the `seven Ws', \ie who, what, where, when, why, how, and which (the `which' questions have not been released at the time of writing this paper). The questions must also relate to the image so as to be clearly answerable if and only if the image is shown. Two types of questions are considered: free-form and region-based. In the free-form setting, the annotator is shown an image and asked to provide 8 question/answer pairs. To encourage diversity, the annotator is forced to use 3 different ``Ws'' out of the 7 mentioned above. In the region-based setting, the annotator must provide questions/answers related to a specific, given region of the image. The diversity of the answers in the Visual Genome is larger than in VQA-real \cite{antol2015vqa}, as shown by the top-1000 most frequent answers only covering about 64\% of the correct answers. In VQA-real, the corresponding top-1000 answers cover as much as 80\% of the test set answers. A major advantage of the Visual Genome dataset for VQA is the potential for using the structured scene annotations, which we examine further in Section~\ref{sec:SGKB}. The use of this information to help designing and training VQA systems is however still an open research question.

The Visual7w \cite{zhu2015visual7w} dataset is a subset of the Visual Genome that contains additional annotations. The questions are evaluated in a multiple-choice setting, each question being provided with 4 candidate answers, of which only one is correct. In addition, all the objects mentioned in the questions are visually grounded, \ie associated with bounding boxes of their depictions in the images.

\renewcommand{\arraystretch}{1.15} %

\rowcolors{1}{}{} %

\begin{table*}
\resizebox{\linewidth}{!}{
\begin{tabular}{l p{4cm}p{4cm}p{4cm}}
\hline \\ [-2.2ex]
DAQUAR \cite{malinowski2014multi} & \includegraphics[height=3cm, width=4cm]{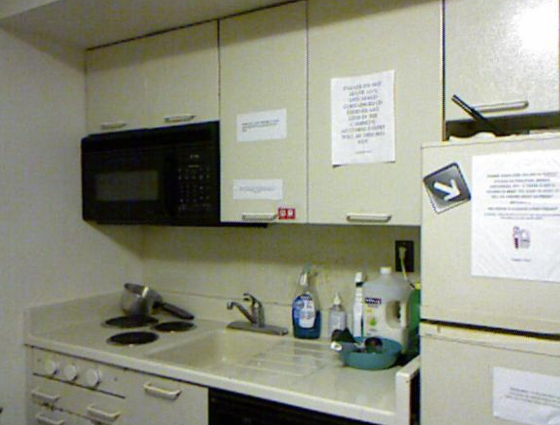} & \includegraphics[height=3cm, width=4cm]{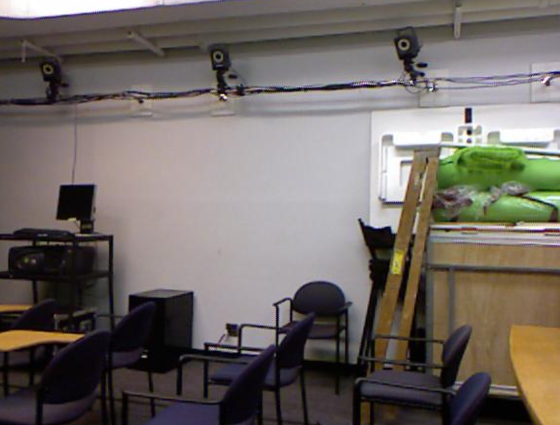}  & \includegraphics[height=3cm, width=4cm]{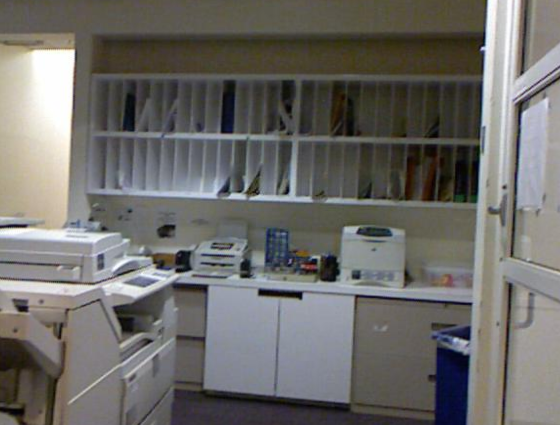}  \\
                      & Q: How many white objects in this picture~? & Q: What color is the chair in front of the wall on the left side of the stacked chairs~? & Q: What is the largest white object on the left side of the picture~? \\
                     & A: 9 & A: blue & A: printer \\ \hline \\ [-2.2ex]
{COCO-QA \cite{ren2015image}}& \includegraphics[height=3cm, width=4cm]{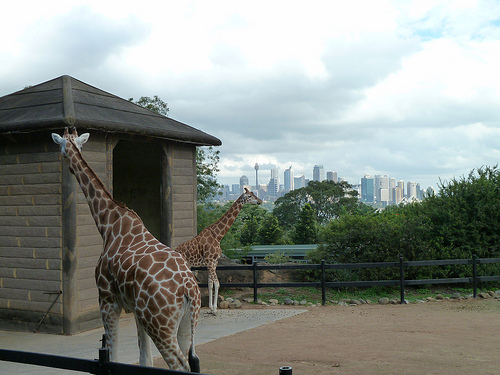} & \includegraphics[height=3cm, width=4cm]{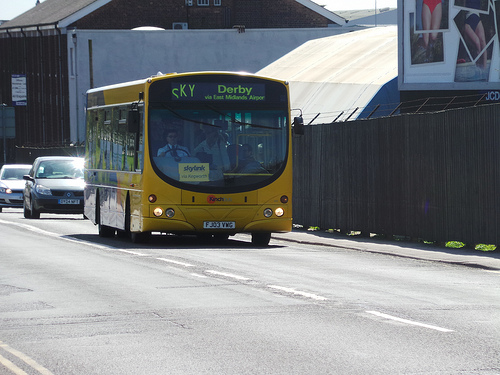}  & \includegraphics[height=3cm, width=4cm]{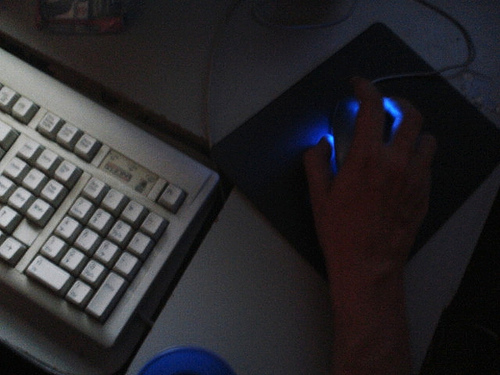}  \\
                      & Q: How many giraffes walking near a hut in an enclosure~? & Q: What is the color of the bus~? & Q: What next to darkened display with telltale blue~? \\
                     & A: two & A: yellow & A: keyboard \\ \hline \\ [-2.2ex]
{VQA-real \cite{antol2015vqa}}& \includegraphics[height=3cm, width=4cm]{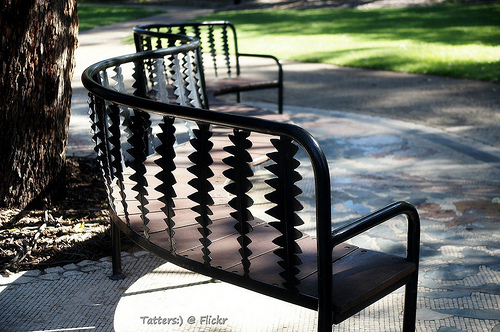} & \includegraphics[height=3cm, width=4cm]{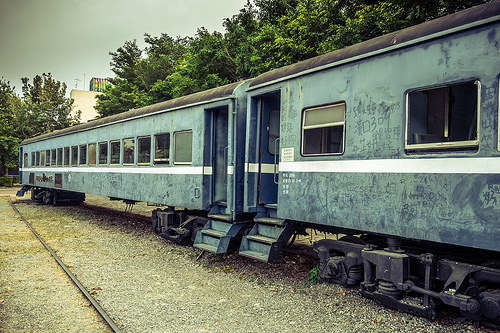}  & \includegraphics[height=3cm, width=4cm]{VQA_3.jpg}  \\
                      & Q: What shape is the bench seat~? & Q: What color is the stripe on the train~? & Q: Where are the magazines in this picture~? \\
                     & A: oval, semi circle, curved, curved, double curve, banana, curved, wavy, twisting, curved & A: white, white, white, white, white, white, white, white, white, white & A: On stool, stool, on stool, on bar stool, on table, stool, on stool, on chair, on bar stool, stool \\ \hline \\ [-2.2ex]
{Visual Genome \cite{krishnavisualgenome}}& \includegraphics[height=3cm, width=4cm]{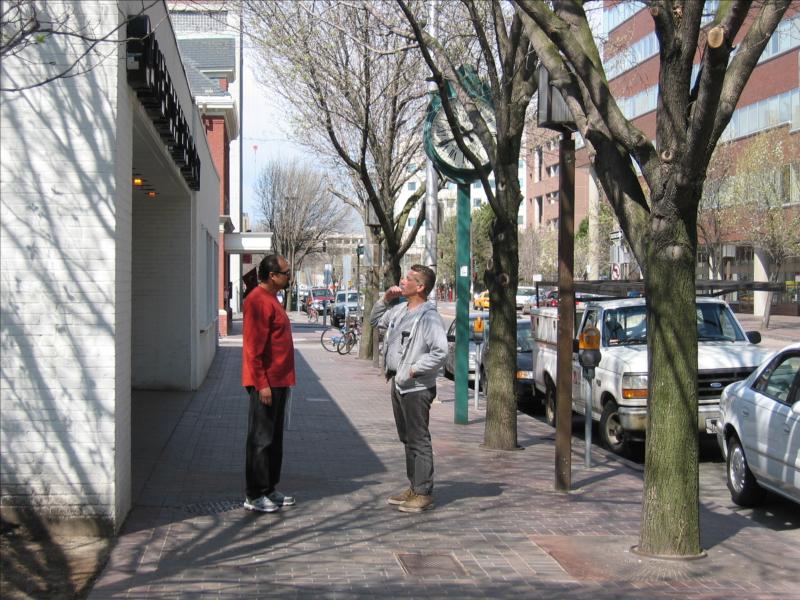} & \includegraphics[height=3cm, width=4cm]{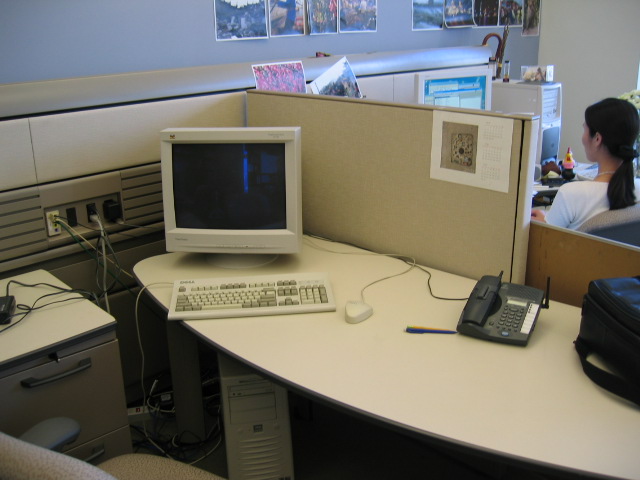}  & \includegraphics[height=3cm, width=4cm]{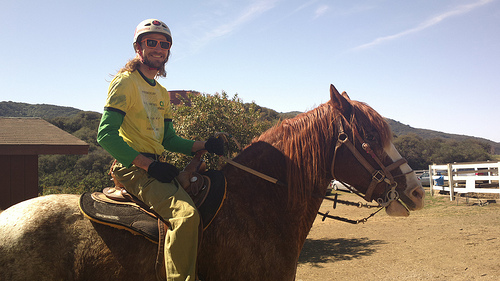}  \\
                      & Q: What color is the clock~? & Q: What is the woman doing~? & Q: How is the ground~? \\
                     & A: Green & A: Sitting & A: dry \\ \hline \\ [-2.2ex]
\end{tabular}}

\caption{Examples from datasets of natural images. The questions in different datasets span a wide range of complexity, involving purely visual attributes and single objects, or more complex relations, actions, and global scene structure. Note that in ``VQA-real'' \cite{antol2015vqa}, every question is provided with 10 answers, each proposed by a different human annotator.}
\label{fig:examples1}
\end{table*}

\begin{table*}
\resizebox{\linewidth}{!}{
\begin{tabular}{l p{4cm}p{4cm}p{4cm}}
\hline \\ [-2.2ex]
{VQA-abstract \cite{antol2015vqa}}& \includegraphics[height=3cm, width=4cm]{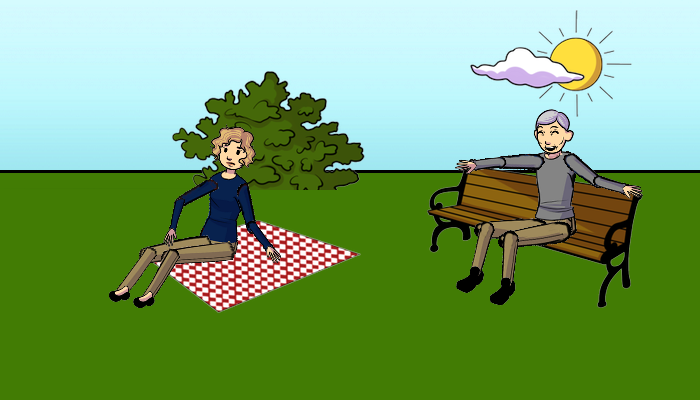} & \includegraphics[height=3cm, width=4cm]{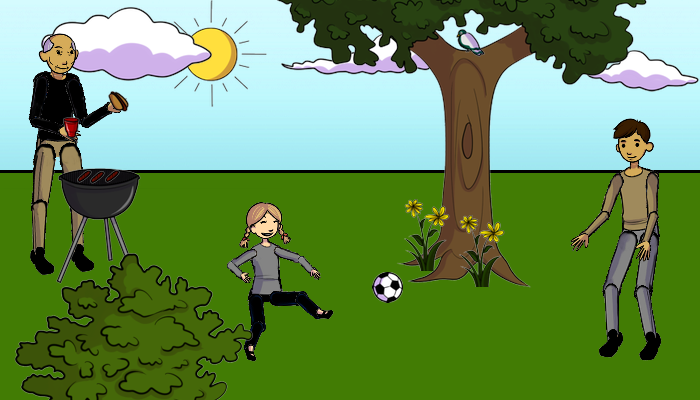}  & \includegraphics[height=3cm, width=4cm]{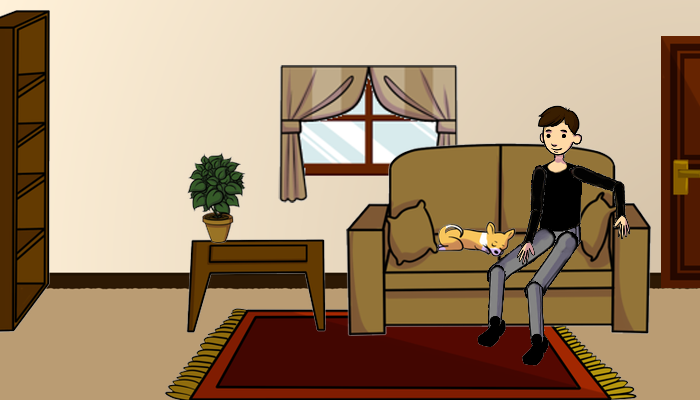}  \\
                      & Q: Who looks happier ?. & Q: Where are the flowers~? & Q: How many pillows~? \\
                     & A: old person, man, man, man, old man, man, man, man, man, grandpa & A: near tree, tree, around tree, tree, by tree, around tree, around tree, grass, beneath tree, base of tree & A: 1, 2, 2, 2, 2, 2, 2, 2, 2, 2 \\ \hline
\end{tabular}}

\caption{Examples from the ``VQA-abstract'' \cite{antol2015vqa} dataset of clip art images. Every question is provided with 10 answers, each proposed by a different human annotator.}
\label{fig:examples2}
\end{table*}

\begin{table*}
\resizebox{\linewidth}{!}{
\begin{tabular}{l p{4cm}p{4cm}p{4cm}}
\hline \\ [-2.2ex]
{KB-VQA \cite{wang2015explicit}}& \includegraphics[height=3cm, width=4cm]{KB_1.jpg} & \includegraphics[height=3cm, width=4cm]{KB_2.jpg}  & \includegraphics[height=3cm, width=4cm]{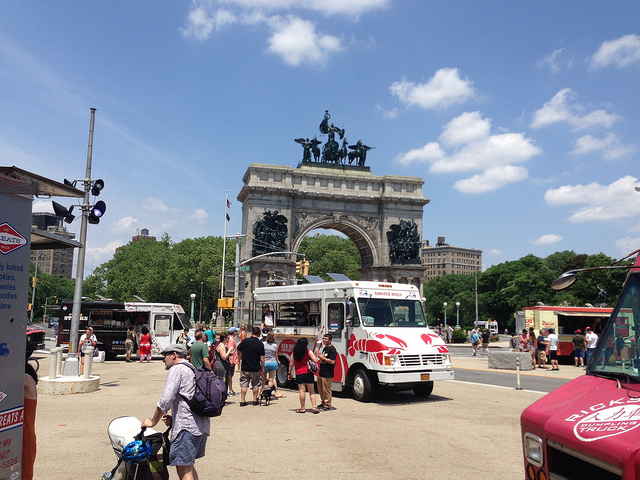}  \\
                      & Q: Tell me the common property of the animal in this image and elephant. & Q: List all equipment I might use to play this sport. & Q: Is the image related to tourism~? \\
                     & A: mammal, animals in Africa & A: baseball bat, baseball, baseball glove, baseball field & A: yes \\ \hline \\ [-2.2ex]
{FVQA \cite{wang2016fvqa}}& \includegraphics[height=3cm, width=4cm]{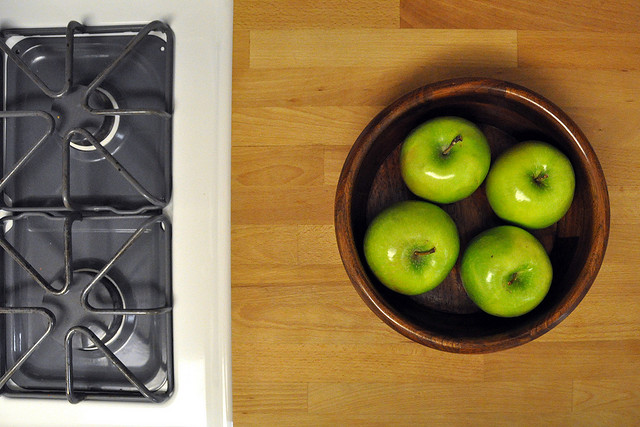} & \includegraphics[height=3cm, width=4cm]{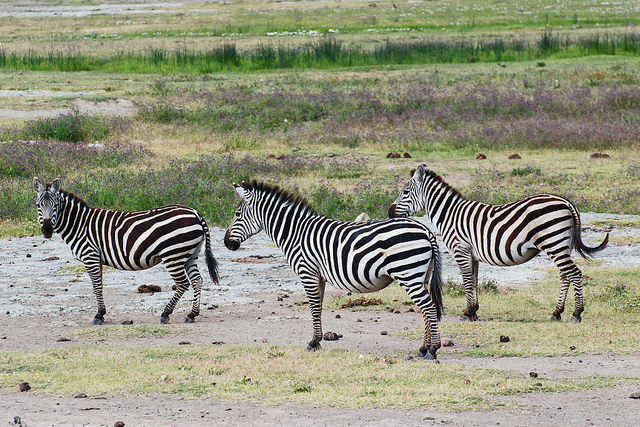}  & \includegraphics[height=3cm, width=4cm]{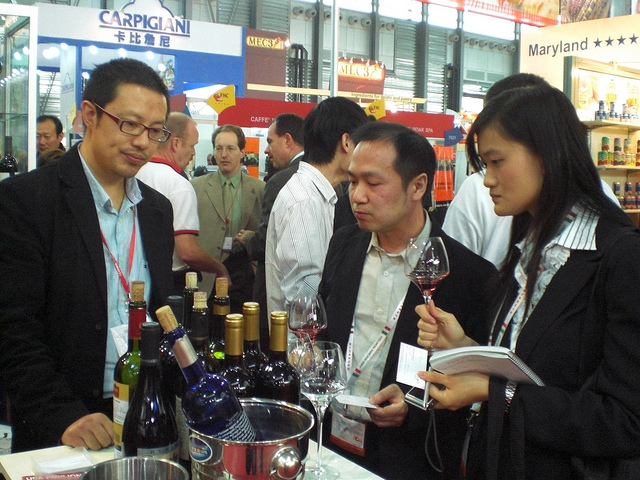}  \\
                      & Q: What things in this image are eatable~?& Q: What is the order of the animal described in this image~? & Q: What thing in this image is helpful for a romantic dinner ? \\
                     & A: Apples & A: Odd toed ungulate & A: Wine \\ \hline                        
\end{tabular}}

\caption{Examples from knowledge base-enhanced datasets.}
\label{fig:examples3}
\end{table*}

\paragraph{Visual Madlibs}
The Visual Madlibs dataset \cite{yu2015visual} is designed to evaluate systems on a ``fill in the blank'' task. The objective is to determine words to complete an affirmation that describes a given image. For example, the description ``two $\rule{1.7em}{.2pt}$ are playing $\rule{1.7em}{.2pt}$ in the park'', provided along a corresponding image, may have to be filled in with ``men'' and ``frisbee''. These sentences essentially amount to questions phrased in declarative form, and most VQA systems could be easily adapted to this setting. The dataset comprises 10,738 images from COCO \cite{lin2014microsoft} and 360,001 focused natural language descriptions. Incomplete sentences were generated automatically from these descriptions using templates. Both open-ended and multiple-choice evaluation are possible.

\paragraph{Evaluation Measures}
The evaluation of computer-generated natural language sentences is an inherently complex task. Both the syntactic (grammatical) and semantic correctness should be taken into account. Comparing generated with ground truth sentences is akin to evaluating paraphrases, which is still an open research problem studied in the NLP community. Most datasets for VQA allow to bypass this issue by restricting answers to single words or short phrases, typically of 1 to 3 words. This allows automatic evaluation, and limits ambiguities during annotation since it forces questions and answers to be more specific.

The seminal paper of Malinowski \etal \cite{malinowski2014multi} proposed two evaluation metrics for VQA. The first is to simply measure the accuracy with respect to the ground truth using string matching. Only exact matches are considered as correct. The second uses the Wu-Palmer similarity (WUPS) \cite{wu1994verbs} which evaluates the similarity between their common subsequence in a taxonomy tree. The candidate answer is considered as correct when the similarity between two words exceeds a threshold. In \cite{malinowski2014multi}, the metric is evaluated against two thresholds, 0.9 and 0.0. In \cite{gao2015you}, Gao \etal conduct an actual \textit{Visual Turing Test} using human judges. Subjects are presented with an image, a question and a candidate answer, from either a VQA system or another human. He or she then needs to determine, based on the answer, whether it was more likely to have been generated by a human (\ie pass the test) or a machine (\ie fail the test). They also rate each candidate answer with a score.

The VQA-real dataset \cite{antol2015vqa} recognize the issue of ambiguous questions and collect, for each question, 10 ground truth answers from 10 different subjects. Evaluation on this dataset must compare a generated answer with these 10 human-generated ones as follows:
\begin{equation}
\text{accuracy} ~=~ \min\Big(\frac{\text{\# humans provided that answer}}{3}, 1\Big)
\end{equation}
In other words, an answer is deemed 100\% accurate if at least 3 annotators provided that exact answer.

\rowcolors{2}{}{gray!25} %
\begin{table}[t!]
\begin{center}
\resizebox{\linewidth}{!}{
    \begin{tabular}{ l c c c }
    \Xhline{2\arrayrulewidth}
    \textbf{DAQUAR-all}  & Acc. (\%)  &  WUPS @0.9  &  WUPS @0.0 \\
    \Xhline{2\arrayrulewidth}
    Neural-Image-QA \cite{malinowski2015ask} & 19.43 & 25.28 & 62.00\\
    Multimodal-CNN \cite{ma2015learning} & 23.40 & 29.59 & 62.95\\
    Attributes-LSTM \cite{wu2015image} &  24.27 &  30.41 &  62.29 \\
    QAM \cite{Chen2015ABC} &  25.37 &  31.35 &  65.89 \\
    DMN+ \cite{xiong2016} & 28.79 & - & -\\
    Bayesian \cite{kafle2016answer} & 28.96 & 34.74 & 67.33\\
    DPPnet \cite{Noh2015image} & 28.98 & 34.80 & 67.81\\
    ACK \cite{wu2015ask}  &  29.16 &  35.30 &  68.66 \\
    ACK-S \cite{wu2016imagepami}  &  29.23 &  35.37 & 68.72\\
    SAN \cite{yang2015stacked} & 29.30 & 35.10 & 68.60\\
    \Xhline{2\arrayrulewidth}
    \end{tabular}}
    \caption{Reported results on the DAQUAR-all dataset.}
    \label{tab:DAQUAR_All}
\end{center}
\end{table}

\rowcolors{2}{}{gray!25} %
\begin{table}[t!]
\begin{center}
\resizebox{\linewidth}{!}{
    \begin{tabular}{ l c c c }
    \Xhline{2\arrayrulewidth}
    \textbf{DAQUAR-reduced}  & Acc. (\%)  &  WUPS @0.9  &  WUPS @0.0 \\
    \Xhline{2\arrayrulewidth}
    GUESS \cite{ren2015image} & 18.24 & 29.65 & 77.59 \\
    VIS+BOW \cite{ren2015image} & 34.17 & 44.99 & 81.48 \\
    VIS+LSTM \cite{ren2015image} & 34.41 & 46.05 & 82.23 \\
    Neural-Image-QA \cite{malinowski2015ask} & 34.68 & 40.76 & 79.54\\
    2-VIS+BLSTM \cite{ren2015image} & 35.78 & 46.83 & 82.15 \\
    Multimodal-CNN~\cite{ma2015learning} & 39.66 & 44.86 & 83.06\\
    SMem \cite{xu2015ask} & 40.07 & - & -\\
    Attributes-LSTM \cite{wu2015image}  &  40.07 &  45.43 &  82.67 \\
    QAM \cite{Chen2015ABC} &  42.76 &  47.62 &  83.04 \\    
    DPPnet~\cite{Noh2015image} & 44.48 & 49.56 & 83.95\\
    Bayesian \cite{kafle2016answer} & 45.17 & 49.74 & 85.13\\
    SAN~\cite{yang2015stacked} & 45.50 & 50.20 & 83.60\\
    ACK \cite{wu2015ask} &  45.79 &  51.53 & 83.91\\
    ACK-S \cite{wu2016imagepami} &  46.13 &  51.83 & 83.95\\
     \Xhline{2\arrayrulewidth}
    \end{tabular}}
    \caption{Reported results on the DAQUAR-reduced dataset.}
    \label{tab:DAQUAR-Reduced}
\end{center}
\end{table}

\rowcolors{2}{}{gray!25} %
\begin{table}[t!]
\begin{center}
\resizebox{\linewidth}{!}{
    \begin{tabular}{ l c c c }
    \Xhline{2\arrayrulewidth}
    \textbf{Toronto COCO-QA}  & Acc. (\%)  &  WUPS @0.9  &  WUPS @0.0 \\
    \Xhline{2\arrayrulewidth}
    GUESS\cite{ren2015image} & 6.65 & 17.42 & 73.44 \\
    VIS+LSTM\cite{ren2015image} & 53.31 & 63.91 & 88.25 \\
    Multimodal-CNN~\cite{ma2015learning} & 54.95 & 65.36 & 88.58\\
    2-VIS+BLSTM\cite{ren2015image} & 55.09 & 65.34 & 88.64 \\
    VIS+BOW\cite{ren2015image} & 55.92 & 66.78 & 88.99 \\
    QAM~\cite{Chen2015ABC} & 58.10 & 68.44 & 89.85\\
    DPPnet~\cite{Noh2015image} & 61.19 & 70.84 & 90.61\\
    Attributes-LSTM \cite{wu2015image} &  61.38 &  71.15 &  91.58 \\
    SAN~\cite{yang2015stacked} & 61.60 & 71.60 & 90.90\\
    Bayesian \cite{kafle2016answer} & 63.18 & 73.14 & 91.32\\
    HieCoAtt \cite{lu2016hierarchical} & 65.40 & 75.10 & 92.00\\
    ACK \cite{wu2015ask} &  69.73 &  77.14 &  92.50 \\
    ACK-S \cite{wu2016imagepami} &  70.98 &  78.35 &  92.87 \\
     \Xhline{2\arrayrulewidth}
    \end{tabular}}
    \caption{Reported results on the COCO-QA dataset.}
    \label{tab:coco_QA}
\end{center}
\end{table}

Other datasets such as \cite{krishnavisualgenome,zhu2015visual7w} simply measure accuracy through the ratio of exact matches between predictions and answers, which is sensible when answers are short and therefore mostly unambiguous. Evaluation in a multiple-choice setting (\eg \cite{yu2015visual}) is straightforward. It makes the task of VQA easier by constraining the output space to a few discrete points, and it eliminates any artifact in the evaluation that could arise from a chosen metric.

\paragraph{Results of existing methods}
Most modern methods for VQA have been evaluated on the VQA-real \cite{antol2015vqa}, DAQUAR \cite{malinowski2014multi}, and COCO-QA \cite{ren2015image} datasets. We summarize results on these three main datasets in Tables~\ref{tab:DAQUAR_All}, \ref{tab:DAQUAR-Reduced}, \ref{tab:coco_QA} and \ref{tab:VQA_result}.

\rowcolors{4}{gray!25}{} %

\begin{table*}[t]
\centering
\resizebox{\linewidth}{!}{
\begin{tabular}{llccccccccccccc}
\Xhline{2\arrayrulewidth}
\hiderowcolors
                                          &    &  \multicolumn{6}{c}{Test-dev}              &    &  \multicolumn{6}{c}{Test-standard}        \\ \cline{3-8} \cline{10-15} 
                                         Method &    &  \multicolumn{4}{c}{Open-ended}  &    &  M.C.    &    &  \multicolumn{4}{c}{Open-ended}  &    &  M.C.   \\ \cline{3-6} \cline{8-8} \cline{10-13} \cline{15-15} 
                                          &    &  Y/N    &  Num.   &  Other   &  All    &    &  All   &    &  Y/N    &  Num.   &  Other   &  All    &    &  All  \\ \cline{1-2} \cline{3-6} \cline{7-8} \cline{9-13} \cline{14-15} 
\Xhline{2\arrayrulewidth}
\showrowcolors
Com-Mem \cite{Jiang2015compositional}     &    &  78.3   &  35.9   &  34.5    &  52.6   &    &  -     &    &  -      &  -      &  -       &  -      &    &  -    \\
Attributes-LSTM \cite{wu2015image}        &    &  79.8   &  36.1   &  43.1    &  55.6   &    &  -     &    &  78.7   &  36.0   &  43.4    &  55.8   &    &  -    \\
iBOWING \cite{zhou2015simple}             &    &  76.5   &  35.0   &  42.6    &  55.7   &    &  -     &    &  76.8   &  35.0   &  42.6    &  55.9   &    &  62.0 \\
Region-Sel \cite{shih2015look}            &    &  -      &  -      &  -       &  -      &    &  62.4  &    &  -      &  -      &  -       &  -      &    &  62.4 \\
DPPnet \cite{Noh2015image}                &    &  80.7   &  37.2   &  41.7    &  57.2   &    &  -     &    &  80.3   &  36.9   &  42.2    &  57.4   &    &  -    \\
VQA team \cite{antol2015vqa}              &    &  80.5   &  36.8   &  43.1    &  57.8   &    &  62.7  &    &  80.6   &  36.5   &  43.7    &  58.2   &    &  63.1 \\
MLP-AQI \cite{jabri2016revisiting}        &    &  -      &  -      &  -       &  -      &    &  -  &    &  -      &  -      &  -       &  -      &    &  65.2 \\
SMem \cite{xu2015ask}                     &    &  80.9   &  37.3   &  43.1    &  58.0   &    &  -     &    &  80.9   &  37.5   &  43.5    &  58.2   &    &  -    \\
Neural-Image-QA \cite{malinowski2015ask}  &    &  78.4   &  36.4   &  46.3    &  58.4   &    &  -     &    &  78.2   &  36.3   &  46.3    &  58.4   &    &  -    \\
NMN \cite{andreas2015deep}                &    &  81.2   &  38.0   &  44.0    &  58.6   &    &  -     &    &  81.2   &  37.7   &  44.0    &  58.7   &    &  -    \\
SAN \cite{yang2015stacked}                &    &  79.3   &  36.6   &  46.1    &  58.7   &    &  -     &    &  -      &  -      &  -       &  58.9   &    &  -    \\
ACK \cite{wu2015ask}                      &    &  81.0   &  38.4   &  45.2    &  59.2   &    &  -     &    &  81.1   &  37.1   &  45.8    &  59.4   &    &  -    \\
DNMN \cite{andreas2016learning}           &    &  81.1   &  38.6   &  45.5    &  59.4   &    &  -     &    &  -      &  -      &  -       &  59.4   &    &  -    \\
FDA \cite{ilievski2016focused}            &    &  81.1   &  36.2   &  45.8    &  59.2   &    &  -     &    &  -      &  -      &  -       &  59.5   &    &  -    \\
ACK-S \cite{wu2016imagepami}              &    &  81.0   &  38.5   &  45.3    &  59.2   &    &        &    &  81.1   &  37.2   &  45.9    &  59.5   &    &  -    \\
Bayesian \cite{kafle2016answer}           &    &  80.5   &  37.5   &  46.7    &  59.6   &    &        &    &  80.3   &  37.8   &  47.6    &  60.1   &    &  -    \\
DMN+ \cite{xiong2016}                     &    &  80.5   &  36.8   &  48.3    &  60.3   &    &  -     &    &  -      &  -      &  -       &  60.4   &    &  -    \\
MCB \cite{fukui2016multimodal}            &    &  81.7   &  36.9   &  49.0    &  61.1   &    &  -     &    &  -      &  -      &  -       &  -      &    &  -    \\
DualNet \cite{saito2016dualnet}           &    &  82.0   &  37.9   &  49.2    &  61.5   &    &  66.7     &    &  81.9      &  37.8      &  49.7       &  61.7      &    &  66.7    \\
MRN \cite{kim2016multimodal}              &    &  82.3   &  39.1   &  48.8    &  61.5   &    &  66.3  &    &  82.4   &  38.2   &  49.4    &  61.8   &    &  66.3 \\
HieCoAtt \cite{lu2016hierarchical}        &    &  79.7   &  38.7   &  51.7    &  61.8   &    &  65.8  &    &  -      &  -      &  -       &  62.1   &    &  66.1 \\
MCB-Att \cite{fukui2016multimodal}        &    &  82.7   &  37.7   &  54.8    &  64.2   &    &  -     &    &  -      &  -      &  -       &  -      &    &  -    \\ 
Joint-Loss \cite{noh2016training}         &    &  81.9   &  39.0   &  53.0    &  63.3   &    &  67.7     &    &  81.7      &  38.2      &  52.8       &  63.2      &    &  67.3    \\
Ensemble of 7 models \cite{fukui2016multimodal}        &    &  83.4   &  39.8   &  58.5    &  66.7   &    &  70.2     &    &  83.2      &  39.5      &  58.0       &  66.5      &    &  70.1    \\
\Xhline{2\arrayrulewidth}
\end{tabular}}
\caption{Reported results on the VQA-real test set in the open-ended and multiple-choice (M.C.) settings.}
\label{tab:VQA_result}
\end{table*}

%% file: dataset_abstract.tex
\subsection{Datasets of clipart images}
\label{subsec:datasetclip}

This section discusses datasets of synthetic images created manually from clipart illustrations. They are often referred to as ``abstract scenes'' \cite{antol2015vqa}, although this denomination is confusing since they supposedly depict \emph{realistic} situations, albeit in minimalistic representations. Such ``cartoon'' images allow studying connections between vision and language by focusing on high-level semantics rather than on the visual recognition. This type of images has been used before for capturing common sense \cite{fouhey2014predicting,lin2015imagination,vedantam2015commonsense}, learning models of interactions between people \cite{antol2014zeroshot}, generating scenes from natural language descriptions \cite{zitnick2013interpretation}, and learning the semantic relevance of visual features \cite{zitnick2013semantics,zitnick2016adopting}.

\paragraph{VQA abstract scenes}
The VQA benchmark (Section \ref{subsec:datasetreal}) contains clipart scenes with questions/answer pairs as a separate and complimentary set to the real images. The aim is to enable research focused on high-level reasoning, removing the need to parse real images. As such, the scenes are provided as structured (XML) descriptions, in addition to the actual images. The scenes were created manually. Annotators were instructed to represent realistic situations through a drag-and-drop interface. Two types of scenes are possible, indoor and outdoor, each allowing a different set of elements, including animals, objects, and humans with adjustable poses. A total of 50,000 scenes were generated, and 3 questions per scene (\ie a total of 150,000 questions) were collected, in a similar manner as for the real images of the VQA dataset (Section \ref{subsec:datasetreal}). Each question was answered by 10 subjects who also provided a confidence score. Questions are labeled with an answer type: ``yes/no'', ``number'', and ``other''. Interestingly, the distribution of question lengths and question types (based on the first four words of the questions) is similar to those of real images. However, the number of unique one-word answers is significantly lower (3,770 \textit{vs} 23,234), reflecting the smaller variations and limited set of objects in the scenes. Ambiguity in the ground truth answers is also lower with abstract scenes, as reflected by a better inter-human agreement (87.5\% \textit{vs} 83.3\%). Results on these abstract scenes have so far only reported in \cite{antol2015vqa} and \cite{zhang2015yin}.

\paragraph{Balanced dataset}
Another version of the dataset discussed above is presented in \cite{zhang2015yin}. Most VQA datasets present strong biases such that a language-only ``blind'' model (\ie using no visual input) can often guess correct answers. This seriously hampers the original objective of VQA of acting as a proxy to evaluate deep image understanding. Synthetic scenes allow better control over the distribution in the dataset. The authors in \cite{zhang2015yin} balance the existing abstract binary VQA dataset (discussed above) with additional complementary scenes so that each question has both ``yes'' and ``no'' answers for two very similar scenes.

As examples on strong biases can be in the VQA dataset \cite{antol2015vqa}, any question starting with ``What sport is'' can be answered correctly with ``tennis'' 41\% of the time. Similarly, ``What color are the'' is answered correctly with ``white'' 23\% of the time \cite{zhang2015yin}. Overall, half of all questions can be answered correctly by a blind neural network, \ie using the question alone. This rises to more than 78\% for the binary questions.

The resulting balanced dataset contains 10,295 and 5,328 pairs of complementary scenes for the training and test set respectively. Evaluation should use the VQA evaluation metric \cite{antol2015vqa}. Results were reported using combinations of balanced and unbalanced training and test sets \cite{zhang2015yin}, of which we summarize the interesting observations. First, when testing on unbalanced data (\ie the setting of prior work), it is better to train on similarly unbalanced, so as to learn and exploit dataset biases (\eg that 69\% of answers are ``yes''). Second, testing on the new balanced data, it is now better to train on similarly balanced data. It forces models to use visual information, being unable to exploit language biases in the training set. In this setting, blind models perform, as expected, close to chance. The particular model evaluated is a method for visual verification that relies on language parsing and a number of hand designed rules. The authors also provide results in an even harder form, where a prediction is considered correct only when the model can answer correctly both versions (with yes and no answers) of a scene. In this setting, a language-only model gives zero performance, and this arguably constitutes one of the most rigorous metrics to quantify actual deep scene understanding.

One criticism of forcing the removal of biases in a dataset supposedly depicting realistic scenes is that it artificially shifts the distribution away from the real world. Statistical biases that appear in datasets reflect those inherent to the world, and it is arguable how much these should enter the learning process of a general VQA system.

%% file: dataset_kb.tex
\subsection{Knowledge base-enhanced datasets}
\label{subsec:datasetkb}

The datasets discussed above contain various ratio of purely visual questions, and questions that require external knowledge. For example, most questions in DAQUAR \cite{malinowski2014multi} are purely visual in nature, referring to colors, numbers, and physical locations of objects. In the COCO-QA dataset \cite{ren2015image}, questions are generated automatically from image captions which describe the major visual elements of the image. In the VQA dataset \cite{antol2015vqa}, 5.5\% of questions require ``adult-level common sense'', but none require ``knowledge base-level'' knowledge \cite{wang2015explicit}. We discussed in Section~\ref{subsec:KB_app} methods for VQA that make use of external knowledge bases. The authors of two such methods \cite{wang2015explicit,wang2016fvqa} proposed two datasets that allow highlighting this particular capability. The scope of these datasets is different than the general-purpose VQA datasets discussed above, and they are also smaller in scale.

\paragraph{KB-VQA}
The KB-VQA dataset~\cite{wang2015explicit} was constructed to evaluate the performance of the Ahab VQA system~\cite{wang2015explicit}. It contains questions requiring topic-specific knowledge that is present in DBpedia. 700 images were selected from the COCO image dataset~\cite{lin2014microsoft} and 3 to 5 question/answer pairs were collected for each, for a total of 2,402 questions. Each question follows one of 23 predefined templates. The questions require different levels of knowledge, from common sense to encyclopedic knowledge.

\paragraph{FVQA}
The FVQA dataset \cite{wang2016fvqa} contains only questions which involve external (non-visual) information. It was designed to include additional annotations to ease the supervised training of methods using knowledge bases. In contrast with most VQA datasets~\cite{antol2015vqa,gao2015you,ren2015image,zhu2015visual7w} which only provide question/answer pairs, FVQA includes, with each question/answer, a supporting fact. These facts are represented as triple \texttt{(arg1,rel,arg2)}. For example, consider the question/answer ``Why are these people wearing yellow jackets ? For Safety''. It will include the supporting fact \texttt{(wearing bright clothes,aids,safety)}. To collect this dataset, a large number of such facts (triples) related to visual concepts were extracted from the knowledge bases DBpedia~\cite{auer2007dbpedia}, ConceptNet~\cite{liu2004conceptnet}, and Webchild~\cite{tandon2014acquiring,tandon2014webchild}. Annotators chose an image and a visual element of the image, and then had to select one of those pre-extracted supporting facts related to the visual concept. They finally had to propose a question/answer that specifically involves the selected supporting fact. The dataset contains 193,005 candidate supporting facts related to 580 visual concepts (234 objects, 205 scenes and 141 attributes) for a total of 4,608 questions.

%% file: dataset_others.tex
\subsection{Other datasets}
\label{subsec:datasetother}

\paragraph{Diagrams}
Kembhavi \etal \cite{kembhavi2016diagram} propose a dataset for VQA on diagrams, named as AI2 Diagrams (AI2D). It comprises more than 5,000 diagrams representing grade school science topics, such as the water cycle and the digestive system. Each diagram is annotated with segmentations and relationships between graphical elements. The dataset includes more than 15,000 multiple-choice questions and answers. In the same paper, the authors propose a method specifically designed to infer correct answers on this dataset. The method builds structured representations, named diagram parse graphs (DPG) with techniques specifically tailored to diagrams, \eg for recognizing arrows or text with OCR. The DPGs are then used to infer correct answers. In comparison with VQA on natural images, the visual parsing of diagrams remains challenging and the questions often require a high level of reasoning, which make the task very challenging overall.

\paragraph{Shapes}
\label{subsubsect:shapes}
Andreas \etal \cite{andreas2015deep} propose a dataset of synthetic images. It is complimentary to datasets of natural image as it provides different challenges, by emphasizing the understanding of spatial and logical relations among multiple objects. The dataset consists of complex questions about arrangements of colored shapes. The questions are built around compositions of concepts and relations, \eg \textit{Is there a red shape above a circle ?} or \textit{Is a red shape blue ?}. This allowed the authors to highlight the capabilities of the Neural Module Networks (see Section \ref{subsubsec:modules}). Questions contain between two and four attributes, object types, or relationships. There are 244 questions and 15,616 images in total, with all questions having a yes and no answer (and corresponding supporting image). This eliminates the risk of learning biases, as discussed in Section~\ref{subsec:datasetclip}. The authors provide results of their own method and of a reimplementation of a joint embedding baseline \cite{ren2015image}. No other results have been reported so far. Note that this dataset is similar in spirit to the synthetic ``bAbI'' dataset used in textual QA~\cite{weston2015babi}.

%% file: SGKB.tex
\section{Structured scene annotations for VQA}
\label{sec:SGKB}

\begin{figure}[t]
  \begin{center}
  \includegraphics[width=0.98\linewidth]{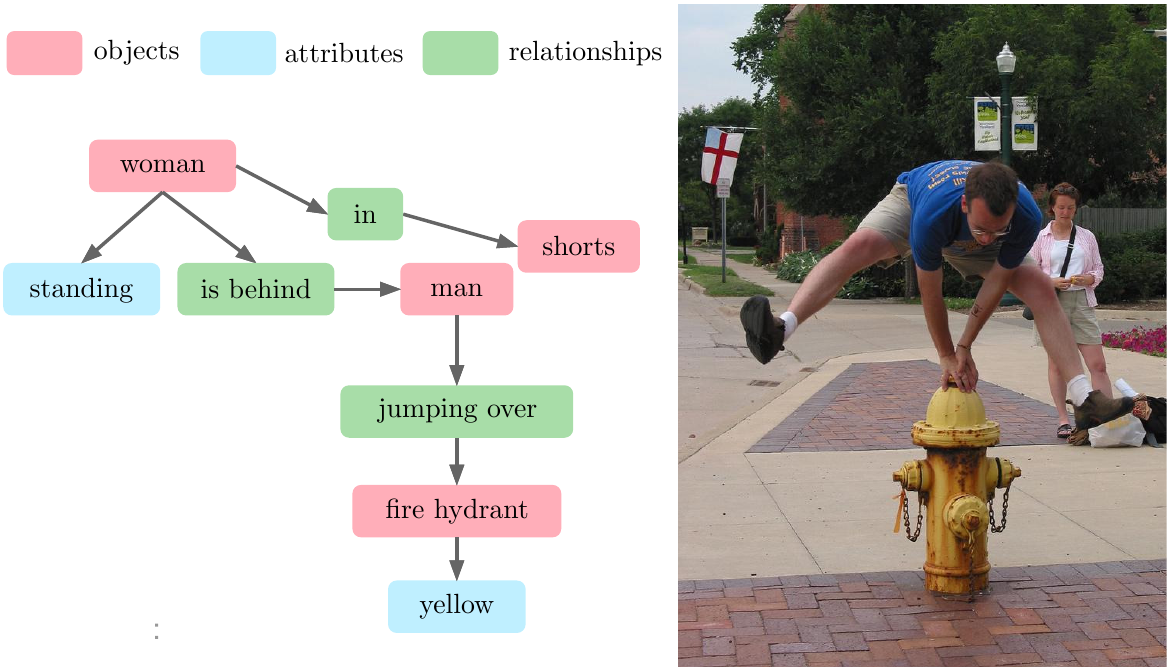}
  \end{center}
  \vspace{-5pt}
  \caption{Example of a scene graph (structured annotation of an image) provided in the Visual Genome dataset \cite{krishnavisualgenome}.}
  \label{fig:scene_example}
\end{figure}

The Visual Genome \cite{krishnavisualgenome} is currently the largest dataset available for VQA. It provides the unique advantage of human-generated structured annotations for each image in the form of scene graphs. In summary, a scene graph is formed of nodes representing visual elements of the scene, which can be objects, attributes, actions, \etc. Nodes are linked with directed edges that represent relationships between them (see Figure~\ref{fig:scene_example} for an example). A detailed description is available in \cite{krishnavisualgenome}.

\begin{figure}[t]
  \begin{center}
  \includegraphics[width=0.98\linewidth]{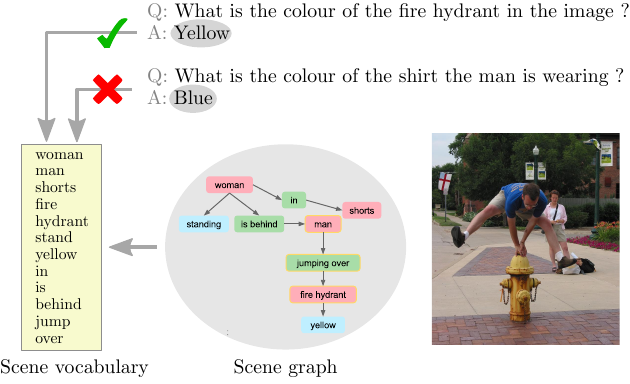}
  \end{center}
  \vspace{-5pt}
  \caption{We verify whether the answer to each question in the Visual Genome dataset can be found from the corresponding scene graph of the scene. Therefore, we first build the vocabulary of each image from the labels of all nodes and edges of its scene graph. Then, for each question, we check whether its answer can be found within the words or combination of words in the vocabulary of the corresponding image.}
  \label{fig:scene_process}
\end{figure}

The inclusion of scene graphs with images is a significant step toward rich and more comprehensive annotations, compared to the more typical object-level and image-level annotations. In this section, we investigate whether scene graphs could be used to directly answer related visual questions. In other words, if we assume a perfect vision system capable of recovering the same scene graph of the input image as the one annotated by a human, could the answer be trivially obtained from it ? We check a prerequisite for such a hypothesis which is whether the answer actually appears as an element of the graph. Practically, we first build a vocabulary for each image based on its corresponding scene graph. Words in the vocabulary are formed from all node labels of the graph. Then, for each question, we check whether its answer can be matched with words or the combination of words from the vocabulary of its image (Figure~\ref{fig:scene_process}) 

We apply the above procedure on all images and questions of the Visual Genome dataset. We find that only 40.02\% of the answers can be directly found in the scene graph, \ie only 40.02\% of the questions could be directly answered using the scene graph representation. Another 7\% of answers are numbers (\ie counting questions) which we choose to leave aside from the rest of our analysis. There remains 53\% of questions can not be directly answered from the scene graph. This ratio is surprisingly high considering the apparent level of detail of the descriptions provided as scene graphs.

\begin{figure}[t]
  \begin{center}
  \includegraphics[width=0.98\linewidth]{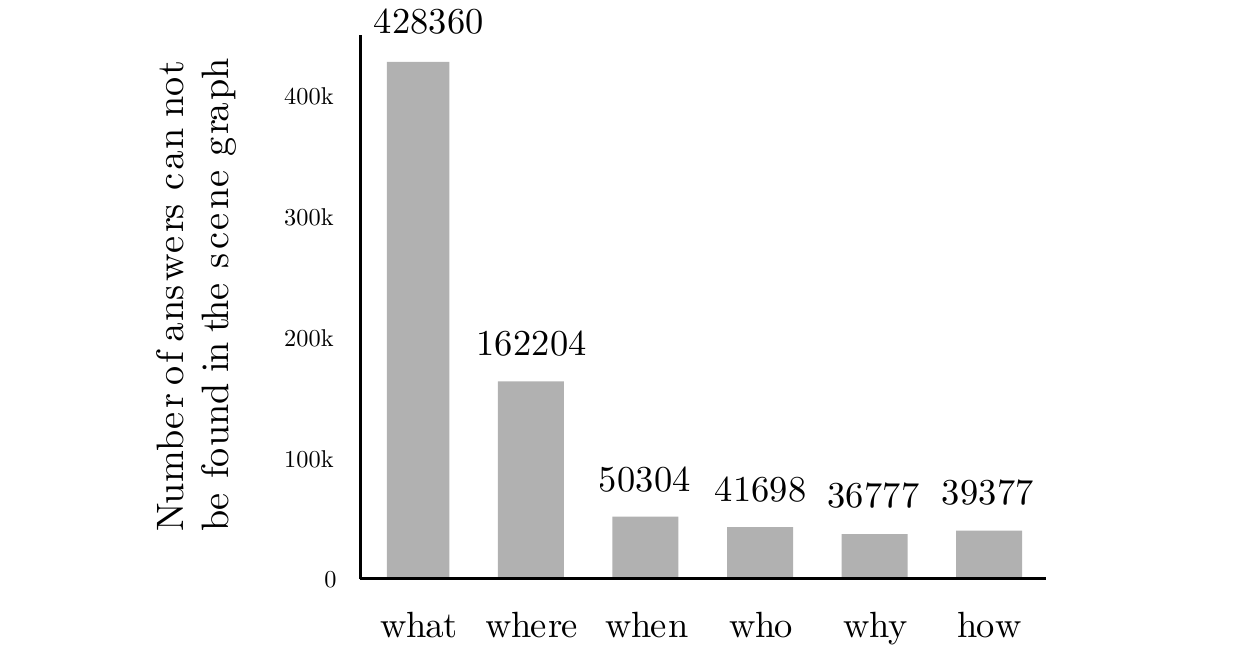}
  \end{center}
  \vspace{-5pt}
  \caption{Number of answers that cannot be found in the scene graph for each question type.}
  \label{fig:scene_bar}
\end{figure}

\begin{figure}[t]
  \begin{center}
  \includegraphics[width=0.98\linewidth]{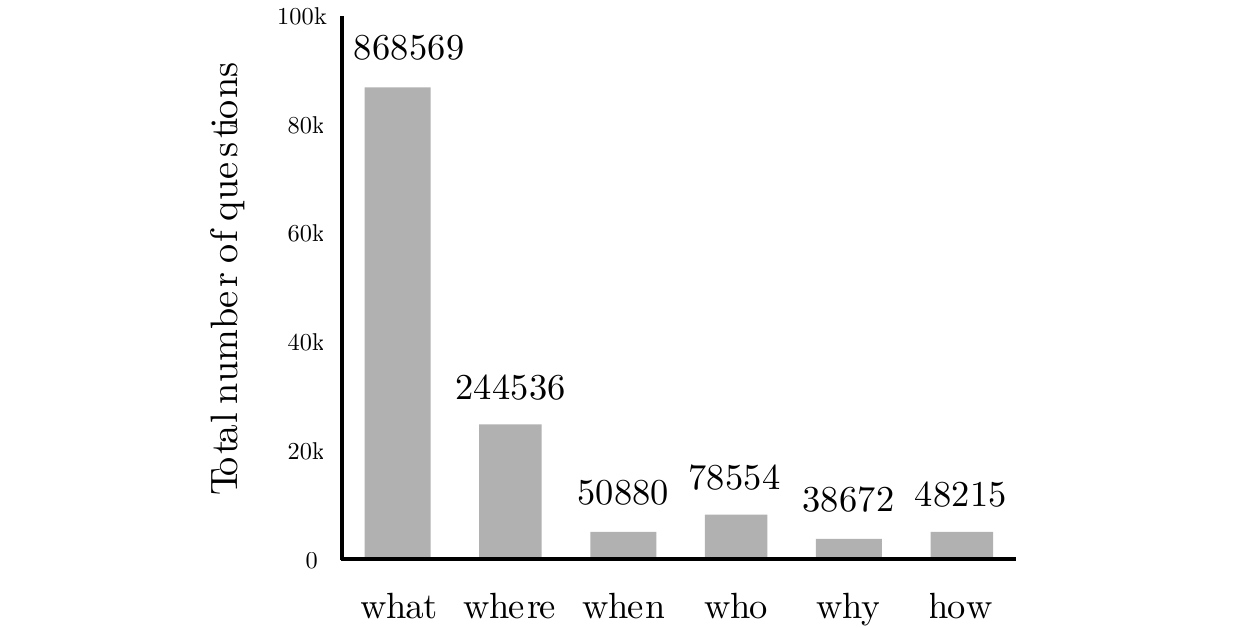}
  \end{center}
  \vspace{-5pt}
  \caption{Total number of questions of each type.}
  \label{fig:scene_bar_2}
\end{figure}

\begin{figure}[t]
  \begin{center}
  \includegraphics[width=0.98\linewidth]{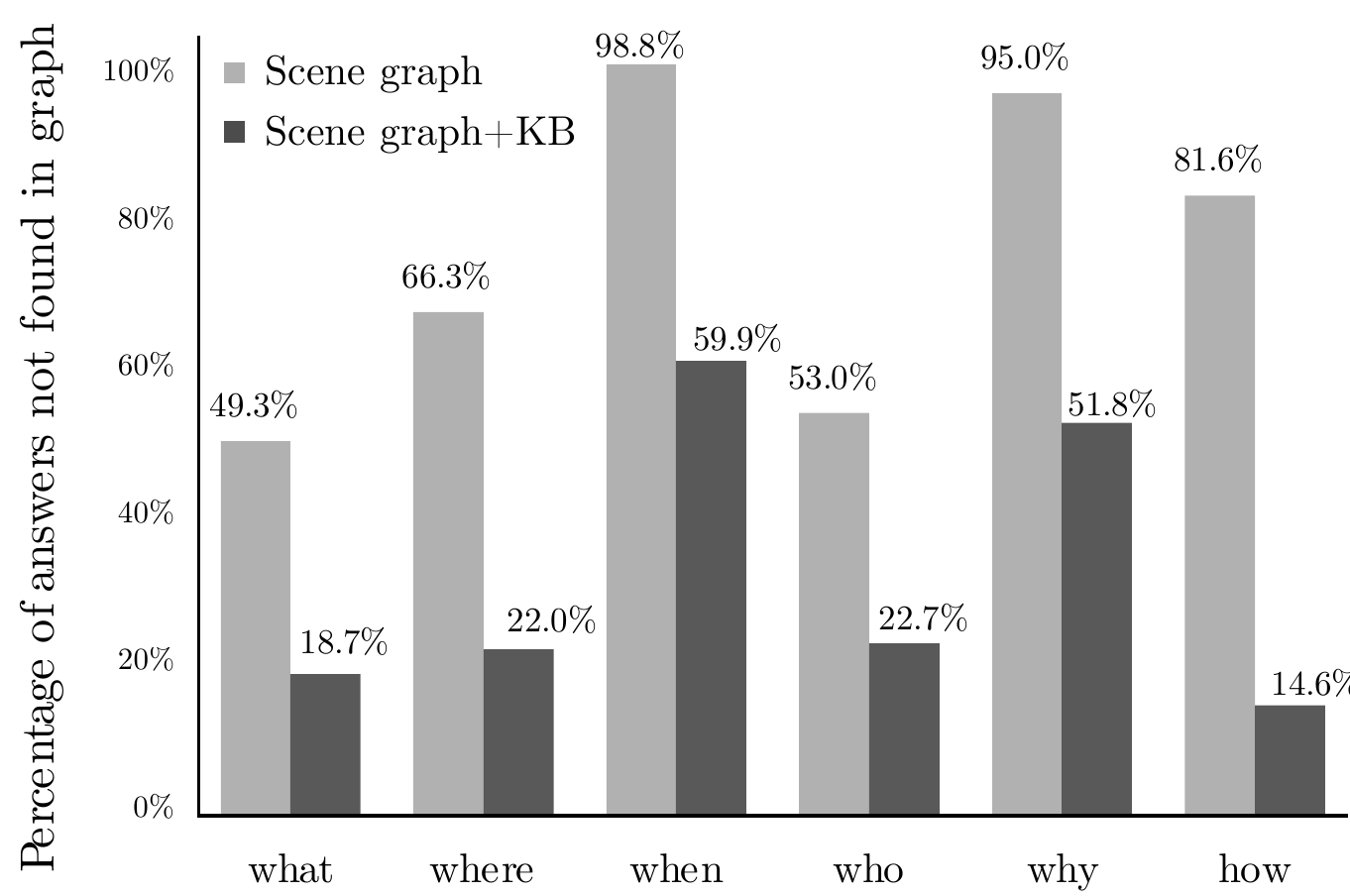}
  \end{center}
  \vspace{-5pt}
  \caption{Answers that can not be found in the scene graph or the knowledge-expanded scene graph, measured as a fraction of all questions of each type separately.}
  \label{fig:scene_bar_3}
\end{figure}

To characterize the remaining 53\% of questions, we examine question types using their first few words (Figure~\ref{fig:scene_bar}. A large number of questions starting with ``what'' are among those that cannot be directly answered by scene graphs. Note that the overall distribution of question types over the whole dataset (including those that could be answered directly from the scene graph) differs significantly (Figure~\ref{fig:scene_bar_2}). We report in Figure~\ref{fig:scene_bar_3} the number of questions that cannot be answered from the scene graph as a fraction of all questions of each type separately. We find that a large fraction of the questions starting with ``when'', ``why'' and ``how'' have answers not be found in scene graphs. Indeed, answering such questions often involves information that does not correspond to specific visual entities, thus not represented by nodes in the scene graphs. It may be possible however to recover these answers using common sense or object-specific knowledge.

We recorded answers that could not be found in scene graphs and ranked them by frequency of occurrence (Table \ref{tab:list}). Answers to ``what'' questions that can not be found in the scene graph are mainly attributes like colours and materials, which are probably considered too fine-grained by annotators for inclusion in the scene graphs. The missing answers to the ``where'' questions are mostly global scene labels such as bedroom, street, zoo, \etc. The missing answers to ``when'' questions are, for 90\% of them, daytime, night, and variants thereof. These labels are seldom represented in the scene graph, and a large number of synonyms can represent a similar semantic concept. Finally, the ``why'' and ``how'' questions lead to higher-level concepts such as actions, reasons, weather types, \etc.

\rowcolors{1}{}{} %

\begin{table*}[t]
\centering
\label{tab:list}
\begin{tabular}{l p{13cm}}
\hline
\textit{what}          & white, green, brown, black, blue, wood, red, gray, yellow, grey, black and white, metal, silver, orange, tree, male, tan, round, sunny, brick, grass, skateboarding, cloud, daytime, surfing, right, skiing, left, pink, dirt, female, standing, water, ...        \\ \hline
\textit{where}         & ground, air, street, table, field, road, sidewalk, zoo, left, sky, right, water, beach, wall, kitchen, background, restaurant, park, bathroom, living room, ocean, in distance, tennis court, plate, airport, bedroom, city, baseball field, ...               \\ \hline
\textit{when}          & daytime, during day, day time, during daytime, in daytime, afternoon, at night, night time, winter, now, nighttime, during day time, night, morning, outside during day time, during daylight hour, evening, daylight, sunny day, daylight hour, ...                 \\ \hline
\textit{who}           & no one, nobody, man, person, woman, photographer, boy, lady, girl, pilot, surfer, child, man on right, man on left, skateboarder, tennis player, no person, little girl, spectator, skier, conductor,guy, passenger, young man, man and woman, ...                  \\ \hline
\textit{why}           & sunny, to hit ball, to eat, sunlight, raining, safety, daytime, to be eaten, during day, remembrance, for fun, for balance, cold, to surf, resting, to play tennis, protection, sun, balance, winter, to catch frisbee, decoration, to play, ...                     \\ \hline
\textit{how}           & clear, none, sunny, cloudy, overcast, calm, open, good, closed, white, clear blue, happy, standing, short, partly cloudy, rainy, green, blue, cold, wet, black, brown, in motion, long, down, blurry, dirty, small, up, large, clean, gray, upside down, sliced, ... \\ \hline
\end{tabular}
\caption{Frequent answers that cannot be found in scene graphs of the input image, for each question type, ranked by rate of occurrence.}
\end{table*}

The above analysis leads to the conclusion that the current scene graphs are rich intermediate abstractions of the scenes, but they are not comprehensive enough to capture \emph{all} elements required for VQA. For example, low-level visual attributes such as colour and material are lost in this representation and one therefore needs to access the image to answer questions involving them. Global scene attributes such as location, weather, or time of day are seldom labeled but they are often involved in ``where'' and ``when'' questions. It remains debatable whether more human effort should be invested to obtain comprehensive annotations. Another solution is to combine visual datasets with large-scale knowledge bases (KBs) that provide common sense information about visual and non-visual concepts. The type of information in those KBs is complementary to visual annotations like scene graphs. For example, although ``daytime'' may not be annotated for a particular scene, the annotation of a ``clear blue sky'' may lead to reason about daytime given some common sense knowledge. Similar reasoning could \eg associate ``oven'' to ``kitchen'', ``food'' to ``eat'', and ``snow'' to ``cold''.

We perform an additional experiment to estimate the potential of connecting scene graphs with a general purpose KB.

For each question, we examine whether the correct answer can be found in a first-order extension of the scene graph using relations from relations in the KB.

More specifically, we use the labels of all nodes of the scene graph to query 3 large KBs (DBpedia~\cite{auer2007dbpedia}, WebChild~\cite{tandon2014acquiring,tandon2014webchild}, and ConceptNet~\cite{liu2004conceptnet}). The triples resulting from the query are used to expand the scene graph and its vocabulary. For example, a scene graph node labeled ``cat'' may return the fact \texttt{<cat,isa,mammal>}, which will be appended to the ``cat'' node of the scene graph. This simple procedure effectively completes the scene-specific graph with general, relevant knowledge. Similarly as above, we then examine whether questions could potentially be answered from this representation alone, checking for matches between the correct answer and words or combination of words from the expanded vocabulary. We find that this is the case for 79.58\% of the questions, which is nearly the double of the same experiment without the KB expansion (40.02\%). This clearly shows the potential for complementing the interpretation of visual contents with information from general-purpose KBs.

%% file: future.tex
\section{Discussion and future directions}
\label{sec:FD}

The introduction of the task of VQA is relatively recent and it sparked significant interest and accelerating developments in just a few years. VQA is a complex task and it was initially encouraged by a certain level of maturity reached in the fundamental tasks of computer vision such as image recognition. VQA is particularly attractive because it constitutes an AI complete task in its ultimate form, \ie considering open-world free-form questions and answers. Recent results, although encouraging, should however not fool us, as this ultimate goal is indisputably a long way from any current technique. Reduced and limited forms of VQA, \eg multiple-choice format, short answer lengths, limited types of questions, \etc, are reasonable intermediate objectives that seem attainable. Their evaluation is practically easier and may be more representative of our actual progress.

Our review of datasets (Section~\ref{subsec:datasets}) showed a diversity of protocols for collecting data (from human annotators or semi-automatically from image captions) and imposing certain constraints (\eg focusing on certain image regions, objects, or types of questions). These choices influence the collected questions and answers in many ways. First, they impact the level of complexity and number of facts involved, \ie whether the correct answers can be inferred after recognizing a single item/relation/attribute, or requires inference over multiple elements and characteristics of the scene. Second, they influence the ratio of visual \textit{vs} textual understanding required. One extreme example is the synthetic ``Shapes'' dataset (Section \ref{subsubsect:shapes}) which only requires recognizing a handful of shapes and colors by their names, and rather places the emphasis on the  reasoning over relationships between such elements. Third, they influence the amount of prior external knowledge required. External is to be understood in the sense of not inferable from the given visual and textual input. This information may however be visual in nature, \eg bright blue skies occur during daytime, or yellow jackets are usually worn for safety.

\paragraph{External knowledge}
As mentioned above, VQA constitutes an AI-complete challenge since most tasks in AI can be formulated as questions over images. Note however that these questions will often require external knowledge to be answered. This is a reason for the recent interest in methods connecting VQA with structured knowledge bases, and in specific datasets of questions requiring such mechanisms. One may argue that such complex questions are a distraction from the purely visual questions that should be tackled first. We believe that both paths can be explored in parallel. Unfortunately, the current approaches that use knowledge bases for VQA present serious limitations. \cite{wang2015explicit,wang2016fvqa} can only handle a limited number of question types predefined by hand-coded templates. \cite{wu2015ask} encode retrieved information using Doc2Vec, but the encoding process is question-independent and may include information irrelevant to the question. The concept of memory-augmented neural networks could offer a suitable and scalable framework for incorporating and adaptively selecting relevant external knowledge for VQA. This avenue has not been explored yet to our knowledge. On the dataset front, questions involving significant external knowledge are unevenly represented. Specific datasets have been proposed with such questions and additional annotations of supporting facts, but they are limited in scale. Efforts on datasets will likely stimulate research in this direction and help training suitable models. Note that these efforts could simply involve additional annotations of existing datasets.

\paragraph{Textual question answering}
The task of textual question answering predates its visual counterpart by several decades and has produced a substantial amount of work. Two distinct types of approaches have traditionally been proposed: information retrieval, and semantic parsing coupled with knowledge bases. On the one hand, information retrieval approaches use unstructured collections of documents, in which the key words of the question are looked for to identify relevant passages and sentences \cite{jurafsky2000book,kolomiyets2011survey}. A ranking function then sorts these candidates, and the answer is extracted from one or several top matches. This approach can be compared to the basic ``joint embedding  with attention'' method of VQA, where features describing each image region are compared to a representation of the question, identifying the region(s) to focus on, then extracting the answer. A key concept is the prediction of answer type from the question (\eg a colour, a date, a person, \etc) to facilitate the final extraction of the answer from candidate passages. This very concept of answer-type prediction was recently brought back to VQA by Kafle and Kanan \cite{kafle2016answer}. On the other hand, the semantic parsing approaches focus on a better understanding of the question, using more sophisticated language models and parsers to turn the question into structured queries \cite{dong2016logical,iyyer2014paragraphs,singh2014survey,yih2015graph}. These queries can then be executed on domain-specific databases or general purpose structured knowledge bases. A similar process is used in the VQA methods of Wang \etal for querying external knowledge bases \cite{wang2015explicit,wang2016fvqa}.

As we stated before, interest in VQA grew from the maturity of deep learning on tasks of image recognition (of objects, activities, scenes, \etc). Most current work on VQA is therefore built with tools and methods from the computer vision community. Textual question answering has traditionally been addressed in the natural language processing community, with different approaches and algorithms. A number of concepts have permeated from NLP to recent efforts on VQA, for example word embeddings, sentence representations, processing with recurrent neural networks, \etc. Some notable successes are attributable to joint efforts from both fields (\eg \cite{andreas2016learning,xiong2016}). We believe that there still exists potential for better use of concepts from NLP for addressing challenges in VQA. Language models are trainable on large amounts of minimally-labeled text, independently from visual data. They can then be used in the output stage of VQA systems to generate long answers in natural language. Similarly, syntactic parsers may be pre-trained on text alone and be reused for a more principled processing of input questions. The understanding of the question does not have to be trained end-to-end as most VQA systems currently do. The interpretation of text queries into logical representations has been studied in NLP in its own right (\eg \cite{zettlemoyer2007onlinelearning,peng2015languageunderstanding,dong2016logical}).

\section{Conclusion}
This article presented a comprehensive review of the state-of-the-art on visual question answering. We reviewed the most popular approach that maps questions and images to vector representations in a common feature space. We described additional improvements that build up on this concept, namely attention mechanisms, modular and memory-augmented architectures. We reviewed the growing number of datasets available for training and evaluating VQA methods, highlighting differences in the type and difficulty of questions that they include. In addition to a descriptive review, we pinpointed a number of promising directions for future research. In particular, we suggest to scale up the inclusion of additional external knowledge from structured knowledge bases, as well as a continued exploration of the potential of natural language processing tools. We believe that the ongoing and future work on these particular points will benefit the specific task of VQA as well as the general objective of visual scene understanding.